\documentclass{article}

\usepackage[numbers]{natbib}
\usepackage[preprint]{neurips_2022}

\usepackage[dvipsnames]{xcolor}         
\definecolor{linkColor}{rgb}{0.18,0.39,0.62}
\usepackage[utf8]{inputenc} 
\usepackage[T1]{fontenc}    
\usepackage[colorlinks=true,linkcolor=linkColor,citecolor=linkColor,filecolor=linkColor,urlcolor=linkColor]{hyperref}       
\usepackage{url}            
\usepackage{booktabs}       
\usepackage{amsfonts}       
\usepackage{nicefrac}       
\usepackage{microtype}      

\usepackage{graphicx}
\usepackage{arydshln}
\usepackage{booktabs}
\usepackage{multirow}
\usepackage{caption}
\usepackage{subcaption}
\usepackage{makecell}
\usepackage{csquotes}
\usepackage{epigraph}

\RequirePackage{algorithm}
\RequirePackage{algorithmic}

\usepackage{multirow}
\usepackage{amsmath}
\usepackage{capt-of}
\usepackage{tabularx}
\usepackage{epsfig}
\usepackage{amssymb}
\usepackage{amsfonts}
\usepackage{booktabs}
\usepackage{scalerel}
\usepackage[inline]{enumitem}
\usepackage{listings}
\usepackage{varwidth}
\usepackage[export]{adjustbox}
\usepackage{tikz}
\usetikzlibrary{tikzmark}

\usepackage{stmaryrd}
\usepackage{bbm}
\usepackage{wrapfig}
\usepackage{pifont}

\definecolor{deepblue}{rgb}{0,0,0.5}
\definecolor{officeblue}{RGB}{0,102,204}
\definecolor{deepred}{rgb}{0.6,0,0}
\definecolor{deepgreen}{rgb}{0,0.5,0}
\definecolor{mybrickred}{RGB}{182,50,28}

\definecolor{fillcolor}{RGB}{216,217,252}

\usepackage{etoolbox}
\usepackage{framed}

\newif\ifxetexorluatex
\ifxetex
  \xetexorluatextrue
\else
  \ifluatex
    \xetexorluatextrue
  \else
    \xetexorluatexfalse
  \fi
\fi
\newcommand*\quotesize{60} 
\newcommand*{\openquote}
   {\tikz[remember picture,overlay,xshift=-4ex,yshift=-2.5ex]
   \node (OQ) {\fontsize{\quotesize}{\quotesize}\selectfont``};\kern0pt}

\newcommand*{\closequote}[1]
  {\tikz[remember picture,overlay,xshift=4ex,yshift={#1}]
   \node (CQ) {\fontsize{\quotesize}{\quotesize}\selectfont''};}

\colorlet{shadecolor}{white}

\newcommand*\shadedauthorformat{\emph} 

\newcommand*\authoralign[1]{%
  \if#1l
    \def\authorfill{}\def\quotefill{\hfill}
  \else
    \if#1r
      \def\authorfill{\hfill}\def\quotefill{}
    \else
      \if#1c
        \gdef\authorfill{\hfill}\def\quotefill{\hfill}
      \else\typeout{Invalid option}
      \fi
    \fi
  \fi}
\newenvironment{shadequote}[2][l]%
{\authoralign{#1}
\ifblank{#2}
   {\def\shadequoteauthor{}\def\yshift{-2ex}\def\quotefill{\hfill}}
   {\def\shadequoteauthor{\par\authorfill\shadedauthorformat{#2}}\def\yshift{2ex}}
\begin{snugshade}\begin{quote}\openquote}
{\shadequoteauthor\quotefill\closequote{\yshift}\end{quote}\end{snugshade}}

\usepackage{amsmath,amsfonts,bm}

\def\eqref#1{equation~\ref{#1}}

\def\1{\bm{1}}

\DeclareMathAlphabet{\mathsfit}{\encodingdefault}{\sfdefault}{m}{sl}
\SetMathAlphabet{\mathsfit}{bold}{\encodingdefault}{\sfdefault}{bx}{n}

\newcommand\our{\textsc{Kosmos-1}}
\newcommand\ours{\textsc{Kosmos-1 (1.6B)}}
\newcommand\tlm{\textsc{LLM}}
\newcommand\tlms{{language model}}
\newcommand\lait{{language-only instruction tuning}}

\usepackage{pifont}
\definecolor{bluecode}{RGB}{0, 150, 199}

\title{Language Is Not All You Need: Aligning Perception with Language Models}

\author{
\vspace{-0.25in} \\
Shaohan Huang\thanks{~Equal contribution. $\dagger$ Corresponding author.},~~Li Dong\footnotemark[1],~~Wenhui Wang\footnotemark[1],~~Yaru Hao\footnotemark[1],~~Saksham Singhal\footnotemark[1],~~Shuming Ma\footnotemark[1] \\
{Tengchao Lv,}~~{Lei Cui,}~~{Owais Khan Mohammed,}~~{Barun Patra,}~~{Qiang Liu,}~~{Kriti Aggarwal}
 \\
{Zewen Chi,}~~{Johan Bjorck,}~~{Vishrav Chaudhary,}~~{Subhojit Som,}~~{Xia Song,}~~{Furu Wei}$^\dagger$ \\
Microsoft \\
\url{https://github.com/microsoft/unilm} 
\vspace{-0.4cm}
\\}

\date{}

\begin{document}
\maketitle

\vspace{-0.15in}
\begin{figure*}[ht]
\centering
\includegraphics[width=0.95\columnwidth]{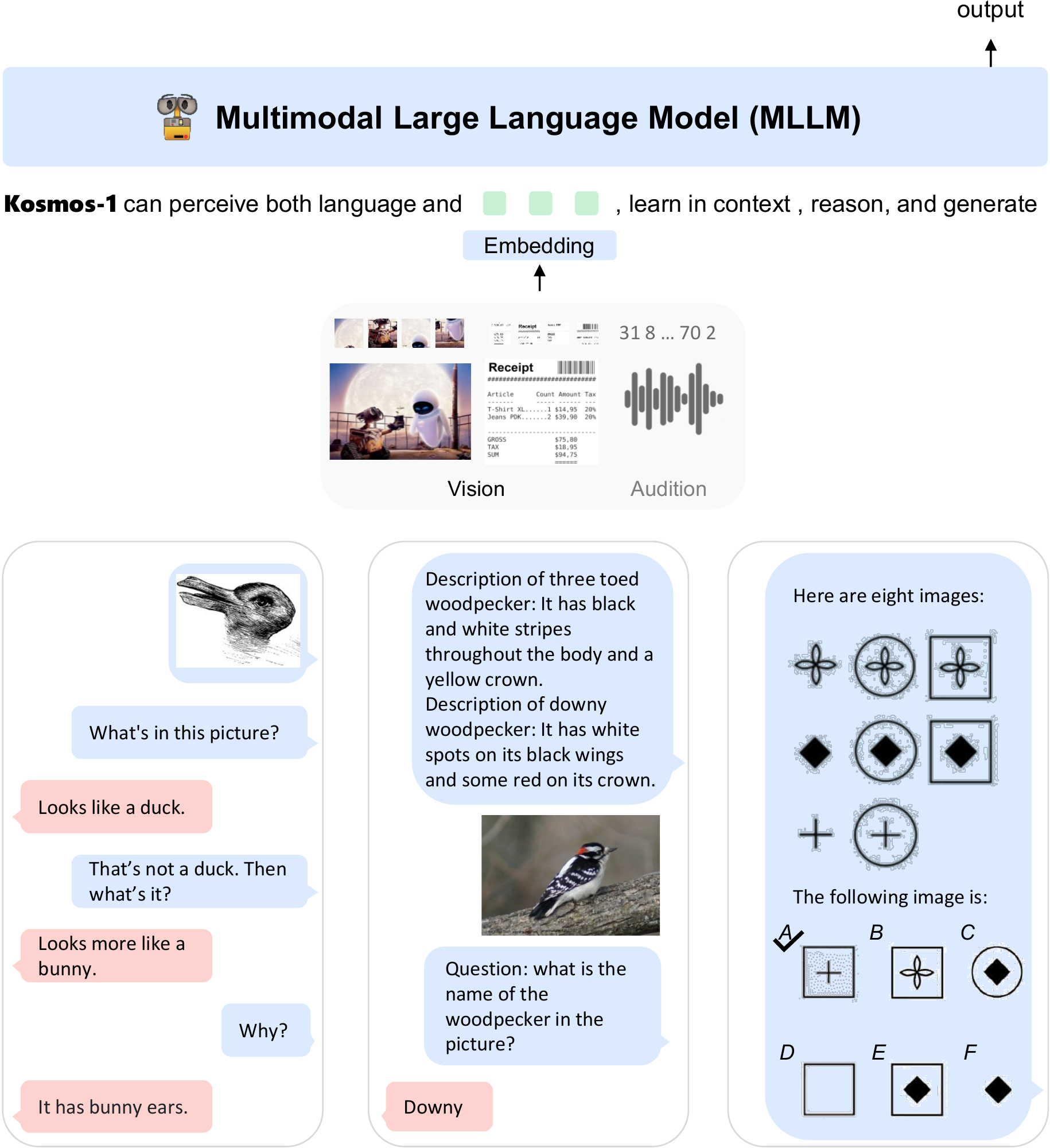}
\caption{
\our{} is a multimodal large language model (MLLM) that is capable of perceiving multimodal input, following instructions, and performing in-context learning for not only language tasks but also multimodal tasks.
In this work, we align vision with large language models (LLMs), advancing the trend of going from LLMs to MLLMs.
}
\label{fig:kosmos}
\end{figure*}

\newpage

\begin{shadequote}[r]{Ludwig Wittgenstein}
The limits of my language means the limits of my world.
\end{shadequote}
\vfill{}

\begin{abstract}
A big convergence of language, multimodal perception, action, and world modeling is a key step toward artificial general intelligence. In this work, we introduce \our{}\footnote{\textsc{Kosmos} is pronounced as and means ``\textit{Cosmos}''.}, a Multimodal Large Language Model ({MLLM}) that can perceive general modalities, learn in context (i.e., few-shot), and follow instructions (i.e., zero-shot). Specifically, we train \our{} from scratch on web-scale multimodal corpora, including arbitrarily interleaved text and images, image-caption pairs, and text data. We evaluate various settings, including zero-shot, few-shot, and multimodal chain-of-thought prompting, on a wide range of tasks without any gradient updates or finetuning. Experimental results show that \our{} achieves impressive performance on (i) language understanding, generation, and even OCR-free NLP (directly fed with document images), (ii) perception-language tasks, including multimodal dialogue, image captioning, visual question answering, and (iii) vision tasks, such as image recognition with descriptions (specifying classification via text instructions). We also show that MLLMs can benefit from cross-modal transfer, i.e., transfer knowledge from language to multimodal, and from multimodal to language. In addition, we introduce a dataset of Raven IQ test, which diagnoses the nonverbal reasoning capability of MLLMs.
\end{abstract}

\vfill{}

\begin{figure*}[ht]
\centering
\includegraphics[width=0.98\textwidth]{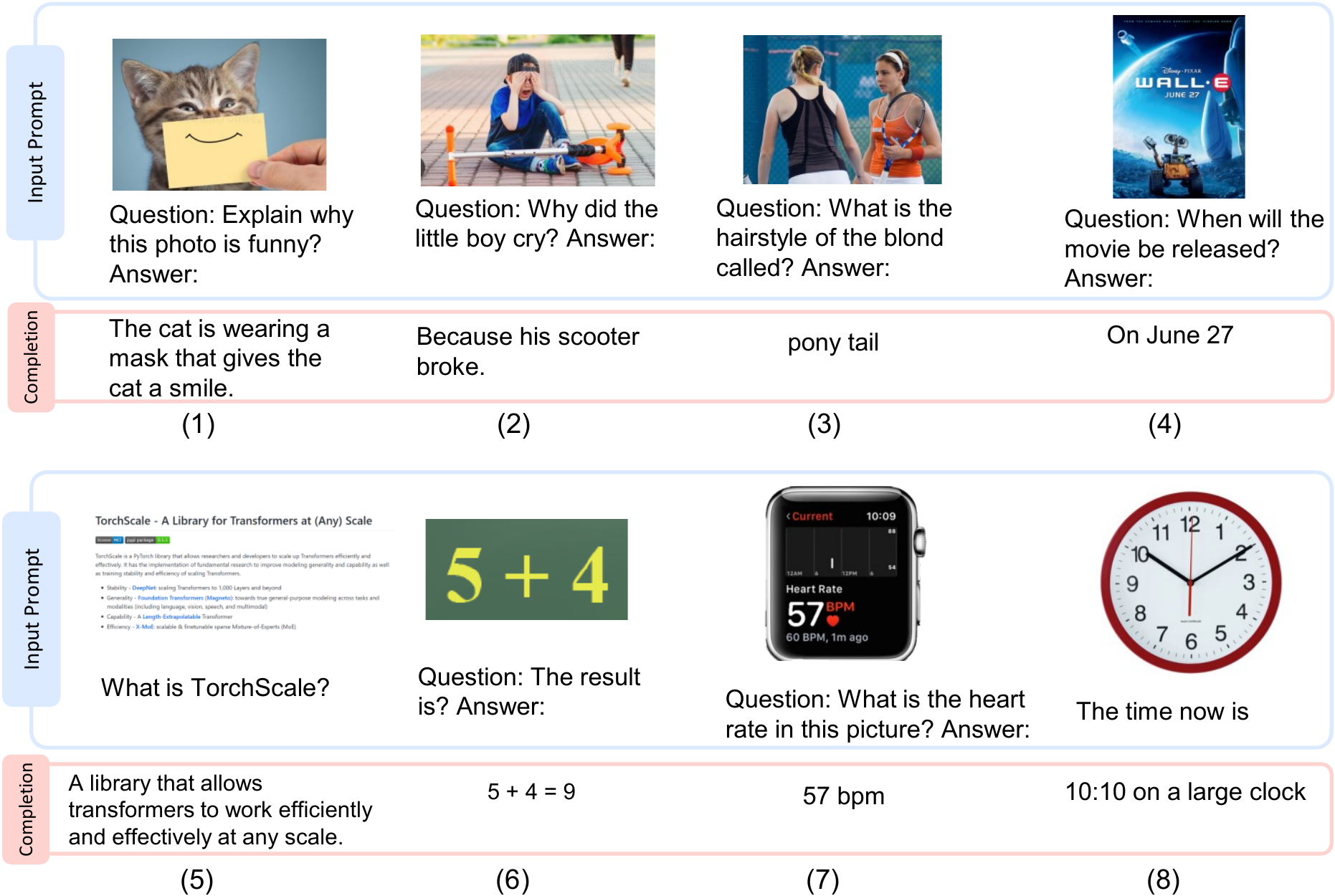}
\caption{
Selected examples generated from \our{}.
Blue boxes are input prompt and pink boxes are \our{} output.
The examples include (1)-(2) visual explanation, (3)-(4) visual question answering, (5) web page question answering, (6) simple math equation, and (7)-(8) number recognition.
}
\label{fig:intro:example:1}
\end{figure*}

\vfill{}

\newpage

\begin{figure*}[ht]
\centering
\includegraphics[width=\textwidth]{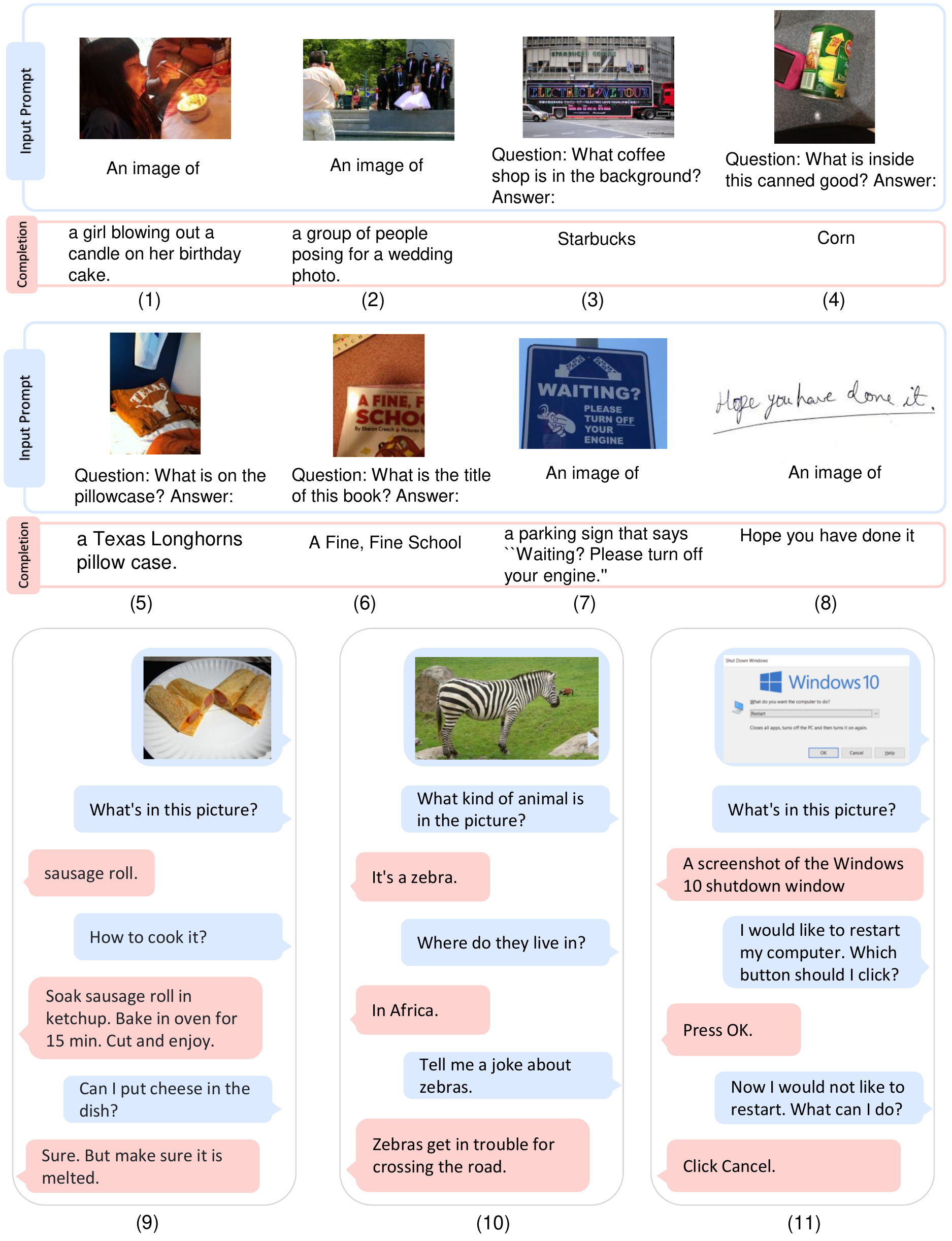}
\caption{
Selected examples generated from \our{}.
Blue boxes are input prompt and pink boxes are \our{} output.
The examples include (1)-(2) image captioning, (3)-(6) visual question answering, (7)-(8) OCR, and (9)-(11) visual dialogue.
}
\label{fig:intro:example:2}
\end{figure*}

\newpage

\section{Introduction: From LLMs to MLLMs}
\label{sec:intro}

Large language models (LLMs) have successfully served as a general-purpose interface across various natural language tasks~\cite{gpt3}.
The LLM-based interface can be adapted to a task as long as we are able to transform the input and output into texts.
For example, the input of the summarization task is a document and the output is its summary. So we can feed the input document into the language model and then produce the generated summary.

Despite the successful applications in natural language processing, it is still struggling to natively use LLMs for multimodal data, such as image, and audio.
Being a basic part of intelligence, multimodal perception is a necessity to achieve artificial general intelligence, in terms of knowledge acquisition and grounding to the real world.
More importantly, unlocking multimodal input~\cite{frozen,metalm,beit3,flamingo,cm3,blip2} greatly widens the applications of language models to more high-value areas, such as multimodal machine learning, document intelligence, and robotics.

In this work, we introduce \our{}, a Multimodal Large Language Model (MLLM) that can perceive general modalities, follow instructions (i.e., zero-shot learning), and learn in context (i.e., few-shot learning).
The goal is to align perception with LLMs, so that the models are able to see and talk.
To be specific, we follow \textsc{MetaLM}~\cite{metalm} to train the \our{} model from scratch.
As shown in Figure~\ref{fig:kosmos}, a Transformer-based language model is regarded as the general-purpose interface, and perception modules are docked with the language model.
We train the model on web-scale multimodal corpora, i.e., text data, arbitrarily interleaved images and texts, and image-caption pairs.
In addition, we calibrate the instruction-following capability across modalities by transferring language-only data.

\begin{table}[t]
\centering
\resizebox{\textwidth}{!}{
\begin{tabular}{@{}lllcc@{}}
\toprule
\textbf{Dataset} & \textbf{Task description}    & \textbf{Metric}                     & \textbf{Zero-shot}    & \textbf{Few-shot} \\
\midrule
\multicolumn{5}{l}{~\textit{Language tasks}} \\
StoryCloze~\cite{storycloze}      &  Commonsense reasoning        & Accuracy         & \ding{51}             & \ding{51}  \\
HellaSwag~\cite{hellaswag}        &  Commonsense NLI              & Accuracy         & \ding{51}             & \ding{51}  \\    
Winograd~\cite{winograd2012}      &  Word ambiguity               & Accuracy         & \ding{51}             & \ding{51}  \\
Winogrande~\cite{winogrande}      &  Word ambiguity               & Accuracy         & \ding{51}             & \ding{51}  \\   
PIQA~\cite{piqa}                  &  Physical commonsense         & Accuracy         & \ding{51}             & \ding{51}  \\         
BoolQ~\cite{boolq}                &  Question answering           & Accuracy         & \ding{51}             & \ding{51}  \\    
CB~\cite{cb}                      &  Textual entailment           & Accuracy         & \ding{51}             & \ding{51}  \\   
COPA~\cite{copa}                  &  Causal reasoning             & Accuracy         & \ding{51}             & \ding{51}  \\ 
Rendered SST-2~\cite{clip}        & OCR-free sentiment classification     & Accuracy          & \ding{51}             &           \\
HatefulMemes~\cite{hateful}       & OCR-free meme classification          & ROC AUC           & \ding{51}             &           \\
\midrule
\multicolumn{5}{l}{~\textit{Cross-modal transfer}} \\
RelativeSize~\cite{Bagherinezhad2016AreEB}     & Commonsense reasoning (object size)    & Accuracy          & \ding{51}             &           \\
MemoryColor~\cite{Norlund2021TransferringKF}   & Commonsense reasoning (object color)     & Accuracy          & \ding{51}             &           \\
ColorTerms~\cite{Bruni2012DistributionalSI}    & Commonsense reasoning (object color)     & Accuracy          & \ding{51}             &           \\
\midrule
\multicolumn{5}{l}{~\textit{Nonverbal reasoning tasks}} \\
IQ Test                            & Raven's Progressive Matrices              & Accuracy          & \ding{51}             &           \\ 
\midrule
\multicolumn{5}{l}{~\textit{Perception-language tasks}} \\
COCO Caption~\cite{mscoco}        & Image captioning             & CIDEr, etc.       & \ding{51}             & \ding{51} \\
Flicker30k~\cite{flickr30k}       & Image captioning             & CIDEr, etc.       & \ding{51}             & \ding{51} \\
VQAv2~\cite{vqav2}                & Visual question answering    & VQA acc.          & \ding{51}             & \ding{51} \\
VizWiz~\cite{vizwiz}              & Visual question answering    & VQA acc.          & \ding{51}             & \ding{51} \\
WebSRC~\cite{websrc}              & Web page question answering  & F1 score          & \ding{51}             &           \\
\midrule
\multicolumn{5}{l}{~\textit{Vision tasks}} \\
ImageNet~\cite{imagenet} & Zero-shot image classification        & Top-1 acc.        & \ding{51}             &           \\
CUB~\cite{CUB:dataset} & Zero-shot image classification with descriptions        & Accuracy   & \ding{51}             &           \\
\bottomrule
\end{tabular}
}
\vspace{0.2cm}
\caption{
We evaluate the capabilities of \our{} on language, perception-language, and vision tasks under both zero- and few-shot learning settings.
}
\label{tab:task-list}
\end{table}

As shown in Table~\ref{tab:task-list}, the \our{} model natively supports language, perception-language, and vision tasks.
We also present some generated examples in Figure~\ref{fig:intro:example:1} and \ref{fig:intro:example:2}.
In addition to various natural language tasks, the \our{} models natively handle a wide range of perception-intensive tasks, spanning visual dialogue, visual explanation, visual question answering, image captioning, simple math equation, OCR, and zero-shot image classification with descriptions.
We also build an IQ test benchmark following Raven's Progressive Matrices~\cite{raven,raven90}, which evaluates the capability of nonverbal reasoning for MLLMs.
The examples show that the native support of multimodal perception enables new opportunities to apply LLMs to new tasks.
Moreover, we show that MLLMs achieve better commonsense reasoning performance compared with LLMs, which indicates cross-modal transfer helps knowledge acquisition.

The key takeaways are as follows:

\paragraph{From LLMs to MLLMs.}
Properly handling perception is a necessary step toward artificial general intelligence.
The capability of perceiving multimodal input is critical to LLMs.
First, multimodal perception enables LLMs to acquire commonsense knowledge beyond text descriptions.
Second, aligning perception with LLMs opens the door to new tasks, such as robotics, and document intelligence.
Third, the capability of perception unifies various APIs, as graphical user interfaces are the most natural and unified way to interact with. For example, MLLMs can directly read the screen or extract numbers from receipts.
We train the \our{} models on web-scale multimodal corpora, which ensures that the model robustly learns from diverse sources.
We not only use a large-scale text corpus but also mine high-quality image-caption pairs and arbitrarily interleaved image and text documents from the web.

\paragraph{Language models as general-purpose interfaces.}
Following the philosophy proposed in \textsc{MetaLM}~\cite{metalm}, we regard language models as a universal task layer.
Because of the open-ended output space, we are able to unify various task predictions as texts.
Moreover, natural-language instructions and action sequences (such as programming language) can be well handled by language models.
LLMs also serve as basic reasoners~\cite{cot}, which is complementary to perception modules on complex tasks.
So it is natural to align world, action, and multimodal perception with the general-purpose interface, i.e., language models.


\paragraph{New capabilities of MLLMs.}

As shown in Table~\ref{tab:task-list}, apart from the capabilities found in previous LLMs~\cite{gpt3,palm}, MLLMs enable new usages and possibilities.
First, we can conduct zero- and few-shot multimodal learning by using natural language instructions and demonstration examples.
Second, we observe promising signals of nonverbal reasoning by evaluating the Raven IQ test, which measures the fluid reasoning ability of humans.
Third, MLLMs naturally support multi-turn interactions for general modalities, such as multimodal dialogue.

\section{\our{}: A Multimodal Large Language Model}
\label{sec:methods}

As shown in Figure~\ref{fig:kosmos}, \our{} is a multimodal language model that can perceive general modalities, follow instructions, learn in context, and generate outputs.
Given the previous context, the model learns to generate texts in an auto-regressive manner.
Specifically, the backbone of \our{} is a Transformer-based causal language model.
Apart from text, other modalities are embedded and fed into the language model.
The Transformer decoder serves as a general-purpose interface to multimodal input.
We train \our{} on multimodal corpora, including monomodal data, cross-modal paired data, and interleaved multimodal data.
Once the models are trained, we can directly evaluate the models in zero-shot and few-shot settings on both language tasks and multimodal tasks.

\subsection{Input Representation}

The Transformer decoder perceives general modalities in a unified way.
For input format, we flatten input as a sequence decorated with special tokens.
Specifically, we use \texttt{<s>} and \texttt{</s>} to denote start- and end-of-sequence. The special tokens \texttt{<image>} and \texttt{</image>} indicate the beginning and end of encoded image embeddings.
For example, ``\texttt{<s>} \textit{document} \texttt{</s>}'' is a text input, and ``\texttt{<s>} \textit{paragraph} \texttt{<image>} Image Embedding \texttt{</image>} \textit{paragraph} \texttt{</s>}'' is an interleaved image-text input.
Table~\ref{tbl:data:format} in Appendix shows some examples of input format.

An embedding module is used to encode both text tokens and other input modalities into vectors. Then the embeddings are fed into the decoder.
For input tokens, we use a lookup table to map them into embeddings.
For the modalities of continuous signals (e.g., image, and audio), it is also feasible to represent inputs as discrete code and then regard them as ``foreign languages''~\cite{beit3,valle}.
In this work, following~\cite{metalm}, we employ a vision encoder as the embedding module for input images.
In addition, Resampler~\cite{flamingo} is used as an attentive pooling mechanism to reduce the number of image embeddings.

\subsection{Multimodal Large Language Models (MLLMs)}

After obtaining the embeddings of an input sequence, we feed them into the Transformer-based decoder.
The left-to-right causal model processes the sequence in an auto-regressive manner, which produces the next token by conditioning on past timesteps.
The causal masking is used to mask out future information.
A $\mathrm{softmax}$ classifier upon Transformer is used to generate tokens over the vocabulary.

MLLMs serve as general-purpose interfaces~\cite{metalm} that can perform interactions with both natural language and multimodal input.
The framework is flexible to handle various data types, as long as we can represent input as vectors.
MLLMs combine the best of two worlds.
First, the language models naturally inherit the capabilities of in-context learning and instruction following.
Second, perception is aligned with language models by training on multimodal corpora.

The implementation is based on the library TorchScale\footnote{\url{https://github.com/microsoft/torchscale}}~\cite{torchscale}, which is designed for large-scale model training.
Compared with the standard Transformer architecture, we include the following modifications:

\paragraph{\textsc{Magneto}}
We use \textsc{Magneto}~\cite{magneto}, a Transformer variant, as the backbone architecture. \textsc{Magneto} has better training stability and superior performance across modalities. It introduces an extra LayerNorm to each sublayer (i.e., multi-head self-attention, and feed-forward network). The method has a theoretically derived initialization method~\cite{deepnet} to improve the optimization fundamentally, which allows us to effectively scale up the models without pain.

\paragraph{\textsc{xPos}}
We employ \textsc{xPos}~\cite{xpos} relative position encoding for better long-context modeling.
The method can better generalize to different lengths, i.e., training on short while testing on longer sequences.
Moreover, \textsc{xPos} optimizes attention resolution so that the position information can be captured more precisely.
The method \textsc{xPos} is efficient and effective in both interpolation and extrapolation settings.

\subsection{Training Objective}

The \our{} training is conducted on web-scale multimodal corpora, including monomodal data (e.g., text corpus), cross-modal paired data (e.g., image-caption pairs), and interleaved multimodal data (e.g., documents of arbitrarily interleaved images and texts).
To be specific, we use monomodal data for representation learning. For example, language modeling with text data pretrains instruction following, in-context learning, and various language tasks.
Moreover, cross-modal pairs and interleaved data learn to align the perception of general modalities with language models.
Interleaved data also naturally fit in the multimodal language modeling task.
We present more details of training data collection in Section~\ref{sec:data}.

The models are trained with the next-token prediction task, i.e., learning to generate the next token depending on the previous context.
The training objective is to maximize the log-likelihood of tokens in examples.
Notice that only discrete tokens, such as text tokens, are accounted for in the training loss.
Multimodal language modeling is a scalable way to train the models. More importantly, the emergence of various capabilities makes the training task favorable for downstream applications.

\section{Model Training}
\label{sec:training}

\subsection{Multimodal Training Data}
\label{sec:data}

The models are trained on web-scale multimodal corpora.
The training datasets consist of text corpora, image-caption pairs, and interleaved data of images and texts.

\paragraph{Text Corpora}

We train our model with The Pile~\cite{pile} and Common Crawl (CC). The Pile is a massive English text dataset built for training large-scale language models, which is produced from a variety of data sources. We exclude data splits from GitHub, arXiv, Stack Exchange, and PubMed Central. We also include the Common Crawl snapshots (2020-50 and 2021-04) datasets, CC-Stories, and RealNews datasets~\cite{megatron,mtnlg}. The entire datasets have been purged of duplicate and near-duplicate documents, as well as filtered to exclude downstream task data. Refer to Appendix~\ref{app:corpora:data:pt:lang} for detailed descriptions of training text corpora.

\paragraph{Image-Caption Pairs}
The image-caption pairs are constructed from several datasets, including English LAION-2B~\cite{laion5b}, LAION-400M~\cite{laion400m}, COYO-700M~\cite{coyo700m}, and Conceptual Captions~\cite{cc3m,cc12m}.
English LAION-2B, LAION-400M, and COYO-700M are collected from web pages of the Common Crawl web data by extracting image sources and the corresponding alt-text.
Conceptual Captions are also from internet web pages.
More details can be found in Appendix~\ref{app:corpora:data:pt:image-caption}.

\paragraph{Interleaved Image-Text Data}
We collect interleaved multimodal data from the Common Crawl snapshot, which is a publicly available archive of web pages. We use a filtering process to select about 71M web pages from the original 2B web pages in the snapshot. We then extract the text and images from the HTML of each selected web page. For each document, we limit the number of images to five to reduce noise and redundancy. We also randomly discard half of the documents that only have one image to increase the diversity. We provide more details about the data collection process in Appendix~\ref{app:corpora:data:pt:multimodal}. By using this corpus, we enable \our{} to handle interleaved text and image and improve its few-shot ability.


\subsection{Training Setup}

The MLLM component has 24 layers with 2,048 hidden dimensions, 8,192 FFN intermediate size, and 32 attention heads, resulting in about 1.3B parameters. We use Magneto's initialization for optimization stability. For faster convergence, the image representation is obtained from a pretrained CLIP ViT-L/14 model with 1,024 feature dimensions. The images are preprocessed into 224$\times$224 resolution during training. We freeze the parameters of the CLIP model except for the last layer during training.
The total number of parameters of \our{} is about 1.6B. More details about hyperparameters can be found in Appendix~\ref{app:hyperparam}.

We use a batch size of 1.2 million tokens (0.5 million tokens from text corpora, 0.5 million tokens from image-caption pairs, and 0.2 million tokens from interleaved data) and train \our{} for 300k steps, corresponding to about 360 billion tokens. We adopt the AdamW optimizer with $\beta=(0.9,0.98)$. We set the weight decay to 0.01 and the dropout rate to 0.1. The learning rate increases to 2e-4 for the first 375 warming-up steps and decays linearly to 0 for the rest of the training steps.
We use SentencePiece~\cite{sentencepiece} to tokenize the text. We preprocess the data in the ``full-sentence'' format~\cite{roberta}, which packs each input sequence with full sentences that are sampled continuously from one or more documents.


\subsection{Language-Only Instruction Tuning}

In order to better align \our{} with human instructions, we perform language-only instruction tuning~\cite{flan2,unnatural}.
Specifically, we continue-train the model with the instruction data in the format of (instructions, inputs, and outputs).
The instruction data is language-only, which is mixed with training corpora.
The tuning process is conducted as language modeling.
Notice that instructions and inputs are not accounted for in the loss.
Section~\ref{sec:eval:instruct} shows that the improvements in the instruction-following capability can transfer across modalities.

We combine Unnatural Instructions~\cite{unnatural} and FLANv2~\cite{flan2} as our instruction dataset. Unnatural Instructions is a dataset that was created by using a large language model to generate instructions for various natural language processing tasks. It has 68,478 instruction-input-output triplets in its core dataset. FLANv2 is a collection of datasets that cover diverse types of language understanding tasks, such as reading comprehension, commonsense reasoning, and closed-book question answering. We randomly select 54k examples of instructions from FLANv2 to augment our instruction dataset.
Details of the training hyperparameter settings are described in Appendix~\ref{app:hyperparam:inst}.


\section{Evaluation}
\label{sec:eval}

MLLMs can handle both language tasks and perception-intensive tasks.
We evaluate \our{} on various types of tasks as follows:

\begin{itemize}
\item Language tasks
\begin{itemize}
 \item Language understanding
 \item Language generation
 \item OCR-free text classification
\end{itemize}
\item Cross-modal transfer
\begin{itemize}
 \item Commonsense reasoning
\end{itemize}
\item Nonverbal reasoning
\begin{itemize}
 \item IQ Test (Raven's Progressive Matrices)
\end{itemize}
\item Perception-language tasks
\begin{itemize}
 \item Image captioning
 \item Visual question answering
 \item Web page question answering
\end{itemize}
\item Vision tasks
\begin{itemize}
 \item Zero-shot image classification
 \item Zero-shot image classification with descriptions
\end{itemize}
\end{itemize}

\subsection{Perception-Language Tasks}
\label{sec:eval:vl}

We evaluate the perception-language capability of \our{} under vision-language settings. Specifically, we conduct zero-shot and few-shot experiments on two widely used tasks, including image captioning and visual question answering. Image captioning involves generating a natural language description of an image, while visual question answering aims to answer a natural language question with respect to an image. 

\subsubsection{Evaluation Setup}

We evaluate the caption generation on MS COCO Caption~\cite{mscoco}, and Flickr30k~\cite{flickr30k}. We use the test set of COCO \textit{Karpathy split}~\cite{karpathysplit}, which re-partitions the train2014 and val2014 images~\cite{mscoco} into 113,287, 5,000, and 5,000 for the training set, validation set, and test set, respectively. We conduct an evaluation on Flickr30k's \textit{Karpathy split} test set. The image resolution is 224$\times$224. We use beam search to generate the captions, and the beam size is 5. In the few-shot settings, we randomly sample demonstrations from the training set. 
We use COCOEvalCap\footnote{\url{https://github.com/salaniz/pycocoevalcap}} to compute CIDEr~\cite{cider} and SPICE~\cite{spice} scores as the evaluation metrics.
We prompt \our{} with \textit{\color{bluecode}``An image of''} for zero-shot and few-shot caption generation experiments.

For visual question-answering tasks, we evaluate zero-shot and few-shot results on test-dev set of VQAv2~\cite{vqav2} and test-dev set of VizWiz~\cite{vizwiz}, respectively. The resolution of images is 224$\times$224. We use greedy search for the decoding.
We follow the normalization rules of the VQAv2 evaluation code\footnote{\url{https://github.com/GT-Vision-Lab/VQA}} when computing the VQA accuracy. 
We evaluate the performance of VQA in an open-ended setting that \our{} generates answers and stops at the \texttt{</s>} (``end of sequence'') token.
 The prompt is \textit{\color{bluecode}``Question: \{question\} Answer: \{answer\}''} for visual question answering tasks.

\subsubsection{Results}

\paragraph{Image Captioning}

Table~\ref{tbl:vl:zs-caption} shows the zero-shot captioning performance on COCO Karpathy test split and Flickr30k test set. 
\our{} achieves remarkable results in zero-shot setting on two image captioning datasets. Specifically, our model achieves a CIDEr score of 67.1 on the Flickr30k dataset, compared to 60.6 and 61.5 for the Flamingo-3B and Flamingo-9B models, respectively. Notably, our model is able to accomplish this feat with a smaller size of 1.6B, compared to Flamingo models. This demonstrates our model's superiority in zero-shot image captioning.

\begin{table}[ht]
\centering
\begin{tabular}{lcccc}
\toprule
\multirow{2}{*}{\textbf{Model}} & \multicolumn{2}{c|}{\textbf{COCO}} & \multicolumn{2}{c}{\textbf{Flickr30k}} \\ \cmidrule(l){2-5} 
 & CIDEr & SPICE & CIDEr & SPICE \\ \midrule
ZeroCap     & 14.6          & 5.5         & -             & -             \\
VLKD        & 58.3          & 13.4        & -             & -             \\
FewVLM      & -             &   -          & 31.0          & 10.0          \\
\textsc{MetaLM}      & 82.2          & 15.7        & 43.4          & 11.7          \\
Flamingo-3B$^*$ & 73.0          & -           & 60.6          & -             \\
Flamingo-9B$^*$ & 79.4          & -           & 61.5          & -             \\
\ours{}      & \textbf{84.7} & \textbf{16.8} & \textbf{67.1} & \textbf{14.5} \\
\bottomrule
\end{tabular}
\vspace{0.2cm}
\caption{Zero-shot image captioning results on COCO caption Karpathy test and Flickr30k test.\\
$^*$ Flamingo~\cite{flamingo} prompts with two examples from the downstream tasks while removing their corresponding images (i.e., similar to few-shot text prompts). The other models do not include any examples in the prompt.
}
\label{tbl:vl:zs-caption}
\end{table}

Table~\ref{tbl:vl:fs-caption} reports the results of the few-shot ($k=2, 4, 8$) settings.
The overall performance improves as the number of shots increases from two to four.
The trends are consistent across the two datasets.
Moreover, the few-shot results outperform zero-shot captioning in Table~\ref{tbl:vl:zs-caption}.

\begin{table}[ht]
\centering
\begin{tabular}{lcccccc}
\toprule
\multirow{2}{*}{\textbf{Model}} & \multicolumn{3}{c|}{\textbf{COCO}} & \multicolumn{3}{c}{\textbf{Flickr30k}} \\ \cmidrule(l){2-7} 
 & $k=2$ & $k=4$ & $k=8$ & $k=2$ & $k=4$ & $k=8$ \\ \midrule
Flamingo-3B & -             & 85.0            & 90.6        & -             & 72.0          & 71.7          \\
Flamingo-9B & -             & 93.1          & \textbf{99.0}          & -             & 72.6          & \textbf{73.4}            \\
\ours{}      & \textbf{99.6} & \textbf{101.7} & 96.7        & \textbf{70.0} & \textbf{75.3} & 68.0          \\ \bottomrule

\end{tabular}
\vspace{0.2cm}
\caption{Few-shot image captioning results on COCO caption Karpathy test and Flickr30k test. CIDEr scores are reported.}
\label{tbl:vl:fs-caption}
\end{table}

\paragraph{Visual Question Answering}

Table~\ref{tbl:vl:zs-qa} reports the zero-shot visual question answering results on VQAv2 and VizWiz. 
We show that \our{} can better handle the diversity and complexity of the VizWiz dataset. \our{} achieves higher accuracy and robustness than Flamingo-3B and Flamingo-9B models.
In addition, our model is competitive with Flamingo on the VQAv2 dataset.

\begin{table}[ht]
\centering
\begin{tabular}{lcc}
\toprule
\textbf{Model} & \textbf{VQAv2}  &  \textbf{VizWiz} \\ \midrule
Frozen       & 29.5          & -             \\
VLKDViT-B/16 & 38.6          & -             \\
\textsc{MetaLM}       & 41.1          & -             \\
Flamingo-3B$^*$  & 49.2          & 28.9          \\
Flamingo-9B$^*$  & \textbf{51.8}          & 28.8          \\
\ours{}      & 51.0          & \textbf{29.2}          \\
\bottomrule
\end{tabular}
\vspace{0.2cm}
\caption{Zero-shot visual question answering results on VQAv2 and VizWiz. We present VQA accuracy scores. ``$*$'': Flamingo~\cite{flamingo} builds the zero-shot prompt with two examples from the downstream tasks where their corresponding images are removed (i.e., similar to few-shot text prompts) while the others evaluate true zero-shot learning.}
\label{tbl:vl:zs-qa}
\end{table}

Table~\ref{tbl:vl:fs-qa} shows the few-shot performance on visual question answering tasks. \our{} outperforms other models in few-shot ($k=2, 4$) settings on the VizWiz dataset. We also observe a positive correlation between the number of shots and the quality of the results on the VizWiz dataset.
Moreover, the few-shot results are better than the zero-shot numbers as reported in Table~\ref{tbl:vl:zs-qa}.

\begin{table}[ht]
\centering
\begin{tabular}{lcccccc}
\toprule
\multirow{2}{*}{\textbf{Model}} & \multicolumn{3}{c|}{\textbf{VQAv2}} & \multicolumn{3}{c}{\textbf{VizWiz}} \\ \cmidrule(l){2-7} 
 & $k=2$ & $k=4$ & $k=8$ & $k=2$ & $k=4$ & $k=8$ \\ \midrule
Frozen      & -             & 38.2          & -             & -              & -             & -             \\
\textsc{MetaLM}      & -             & 45.3          & -             & -              & -             & -             \\
Flamingo-3B & -             & 53.2          & 55.4          & -              & 34.4          & 38.4          \\
Flamingo-9B & -             & \textbf{56.3}          & \textbf{58.0}          & -              & 34.9          & \textbf{39.4}          \\
\ours{}     & \textbf{51.4} & 51.8          & 51.4          & \textbf{31.4}  & \textbf{35.3} & 39.0 \\
\bottomrule
\end{tabular}
\vspace{0.2cm}
\caption{Few-shot visual question answering results on VQAv2 and VizWiz.VQA accuracy scores are reported.
}
\label{tbl:vl:fs-qa}
\end{table}

\subsection{IQ Test: Nonverbal Reasoning}
\label{sec:eval:iq}

\begin{figure*}[t]
\centering
\includegraphics[width=0.98\textwidth]{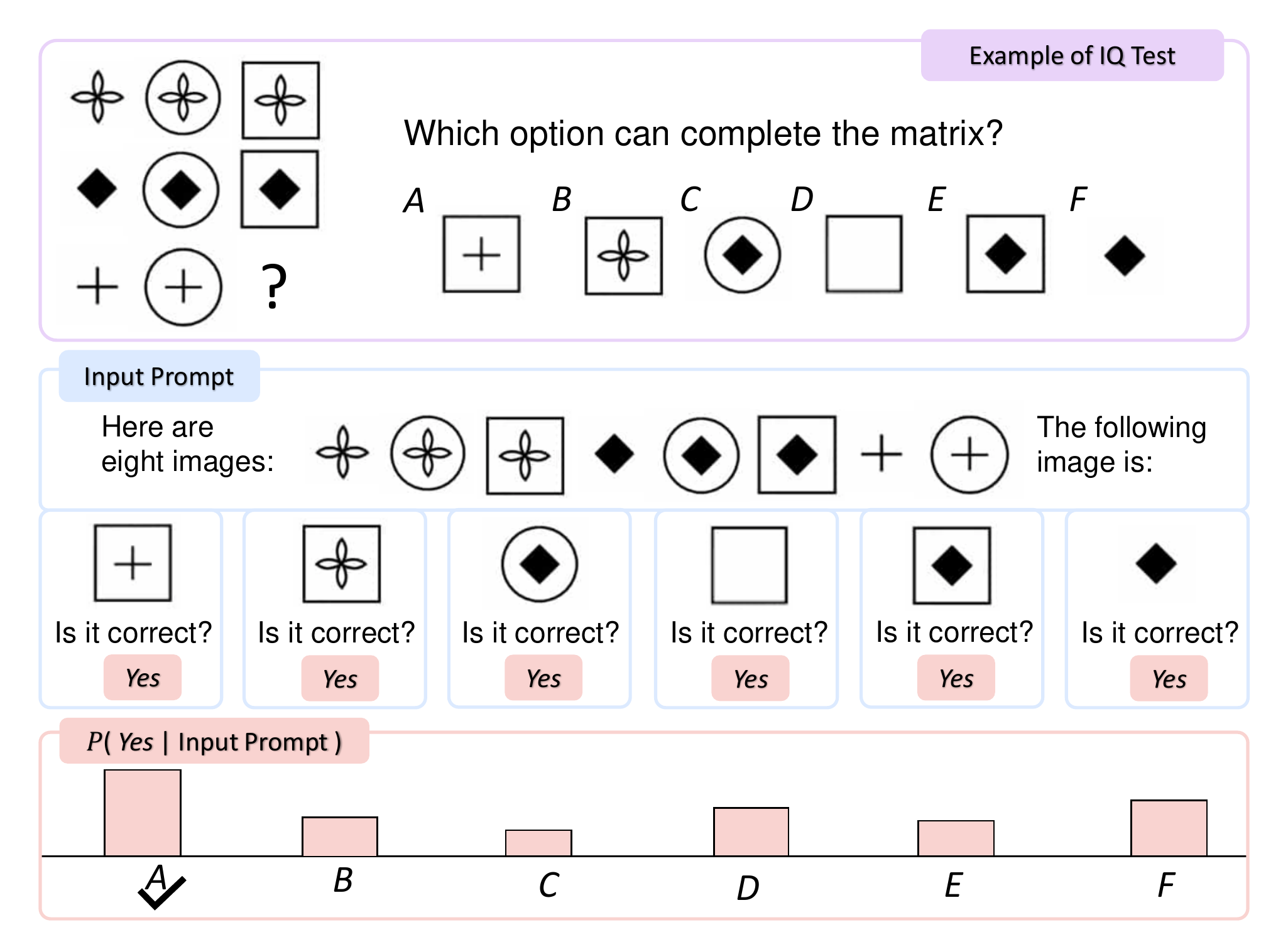}
\caption{\textbf{Top}: An example of Raven IQ test. \textbf{Bottom}: Evaluate \our{} on Raven IQ test. The input prompt consists of the flattened image matrix and verbal instruction. We append each candidate image to the prompt separately and query the model if it is correct.
The final prediction is the candidate that motivates the model to yield the highest probability of ``Yes''.
}
\label{fig:vl:eval:iq}
\end{figure*}

Raven's Progressive Matrices~\cite{raven90,raven} is one of the most common tests to evaluate nonverbal reasoning.
The capability of nonverbal reasoning is typically a reflection of an individual's intelligence quotient (IQ).
Figure~\ref{fig:vl:eval:iq} shows an example. 
Given eight images presented in a $3\times3$ matrix, the task is to identify the following element from six similar candidates.

The models need to conduct zero-shot nonverbal reasoning without explicitly fine-tuning.
The Raven IQ test is analogous to in-context learning of language models, where the difference is whether the context is nonverbal or verbal.
In order to infer the answers, the models have to recognize abstract concepts and identify the underlying patterns of given images.
So the IQ task is a good testbed to benchmark the nonverbal in-context learning capability.

\subsubsection{Evaluation Setup}

To evaluate the \our{} on zero-shot nonverbal reasoning, we construct a dataset of the Raven IQ test. It consists of $50$ examples collected from different websites\footnote{\url{https://en.testometrika.com/intellectual/iq-test/}}\footnote{\url{https://en.testometrika.com/intellectual/iq-test-for-kids-7-to-16-year-old/}}\footnote{\url{https://iqpro.org/}}\footnote{\url{https://iqhaven.com/matrix-g}}.
Each example has three (i.e., $2\times2$ matrix), four, or eight (i.e., $3\times3$ matrix) given images.
The goal is to predict the next one.
Each instance has six candidate images with a unique correct completion.
We measure accuracy scores to evaluate the models.
The evaluation dataset is available at \url{https://aka.ms/kosmos-iq50}.

Figure~\ref{fig:vl:eval:iq} illustrates how to evaluate \our{} on the Raven IQ test.
The matrix-style images are flattened and fed into the models one-by-one.
To enable the model to better understand the desired task, we also use a textual instruction \textit{\color{bluecode}``Here are three/four/eight images:''}, \textit{\color{bluecode}``The following image is:''}, and \textit{\color{bluecode}``Is it correct?''} for conditioning.
We append each possible candidate to the context separately and compare the probability that the model outputs ``Yes'' in a close-ended setting.
The candidate that yields the largest probability is regarded as the prediction.


\subsubsection{Results}

Table~\ref{tbl:vl:iq} shows the evaluation results on the IQ test dataset.
Both \our{} with and without language-only instruction tuning achieve 5.3\% and 9.3\% improvement respectively over the random baseline.
The results indicate that \our{} is able to perceive abstract conceptual patterns in a nonverbal context, and then deduce the following element across multiple choices.
To the best of our knowledge, it is the first time that a model can perform such zero-shot Raven IQ tests.
Although there is still a large performance gap between the current model and the average level of adults, \our{} demonstrates the potential of MLLMs to perform zero-shot nonverbal reasoning by aligning perception with language models.

\begin{table}[ht]
\centering
\begin{tabular}{l c}
\toprule
\bf Method & \bf Accuracy \\
\midrule
{Random Choice} & 17\% \\
{\our{}} & 22\% \\
~ w/o \lait{} & \bf 26\% \\
\bottomrule
\end{tabular}
\vspace{0.2cm}
\caption{Zero-shot generalization on Raven IQ test.}
\label{tbl:vl:iq}
\end{table}

\subsection{OCR-Free Language Understanding}
\label{sec:eval:render}

OCR-free language understanding is a task that focuses on understanding text and images without relying on Optical Character Recognition (OCR). For example, during the Rendered SST-2 task, sentences from the Stanford Sentiment Treebank~\cite{sst} dataset are rendered as images. The model is asked to predict the sentiment of the text within the images.
The task evaluates a model's ability to read and comprehend the meaning of words and sentences directly from the images.

\subsubsection{Evaluation Setup}

We evaluate OCR-free language understanding on the Rendered SST-2~\cite{clip} test set and HatefulMemes~\cite{hateful} validation set. We use accuracy as the metric for the Rendered SST-2 and report ROC AUC for the HatefulMemes dataset. We use the prompt \textit{\color{bluecode}``Question: what is the sentiment of the opinion? Answer: \{answer\}''}, where the answer is either positive or negative for the Rendered SST-2. For the HatefulMemes task, the prompt is  \textit{\color{bluecode}``Question: does this picture contain real hate speech? Answer: \{answer\}''}, where the answer is either yes or no.

\subsubsection{Results}

As shown in Table~\ref{tbl:vl:zero-shot-ocr}, \our{} achieves a ROC AUC of 63.9\% for the HatefulMemes validation set and a test accuracy of 67.1\% for the Rendered SST-2 test set. It outperforms CLIP ViT-L and Flamingo-9B, which achieve AUCs of 63.3\% and 57.0\% on the HatefulMemes task. Note that Flamingo explicitly provides OCR text into the prompt, while \our{} does not access any external tools or resources. This indicates that \our{} has built-in abilities to read and comprehend the text in the rendered images.


\begin{table}[ht]
\centering
\begin{tabular}{l cc}
\toprule
{\textbf{Model}} & {\textbf{HatefulMemes}} & {\textbf{Rendered SST-2}} \\
\midrule
CLIP ViT-B/32 & 57.6          & 59.6          \\
CLIP ViT-B/16 & 61.7          & 59.8          \\
CLIP ViT-L/14 & 63.3          & 64.0          \\
Flamingo-3B   & 53.7          & -             \\
Flamingo-9B   & 57.0          & -             \\
\ours{}        & \textbf{63.9} & \textbf{67.1} \\
\bottomrule
\end{tabular}
\vspace{0.2cm}
\caption{Zero-shot generalization on OCR-free language understanding.  We report accuracy scores.}
\label{tbl:vl:zero-shot-ocr}
\end{table}

\subsection{Web Page Question Answering}
\label{sec:eval:webqa}

Web page question answering aims at finding answers to questions from web pages. 
It requires the model to comprehend both the semantics and the structure of texts. The structure of the web page (such as tables, lists, and HTML layout) plays a key role in how the information is arranged and displayed. The task can help us evaluate our model's ability to understand the semantics and the structure of web pages.

\subsubsection{Evaluation Setup}

We compare the performance on the Web-based Structural Reading Comprehension (WebSRC) dataset~\cite{websrc}.
For comparisons, we train a \tlms{} (\tlm{}) on the same text corpora with the same training setup as in \our{}.
The \tlm{} takes the text extracted from the web page as input. Its template of the prompt is \textit{\color{bluecode}``Given the context below from web page, extract the answer from the given text like this: Qusestion: Who is the publisher of this book? Answer: Penguin Books Ltd. Context: \{WebText\} Q: \{question\} A: \{answer\} ''}, where the \textit{\color{bluecode}\{WebText\}} presents the text extracted from the web page. Besides using the same prompt, \our{} prepends the image before the prompt. Two example images from WebSRC are shown in Appendix~\ref{app:corpora:data:vl:websrc}.
Following the original paper~\cite{websrc}, we use exact match (EM) and F1 scores as our evaluation metrics.

\subsubsection{Results}
The experimental results are summarized in Table~\ref{tbl:vl:websrc}.
We observe that \our{} outperforms the \tlm{}, indicating that \our{} can benefit from the layout and style information of web pages in images.
In addition, we evaluate the performance of \our{} without the extracted text in the prompt. It shows that extracted text has a contribution of +12.0/20.7 EM/F1 to \our{}, indicating that the benefit from modeling images does not sacrifice its language abilities.

\begin{table}[ht]
\centering
\begin{tabular}{lcc}
\toprule
\bf Models & \bf EM & \bf F1 \\
\midrule
\multicolumn{2}{l}{~~\textit{Using extracted text}} \\
\tlm{}              &  7.6        & 17.9 \\
\our{}  &  \textbf{15.8}   & \textbf{31.3} \\
\midrule
\multicolumn{2}{l}{~~\textit{Without using extracted text}} \\
\our{}  &   3.8  & 10.6 \\
\bottomrule
\end{tabular}
\vspace{0.2cm}
\caption{Zero-shot performance on WebSRC task. We report exact match (EM) and F1 scores.}
\label{tbl:vl:websrc}
\end{table}

\subsection{Multimodal Chain-of-Thought Prompting}
\label{sec:eval:cot}

\begin{figure*}[t]
\centering
\includegraphics[width=0.9\textwidth]{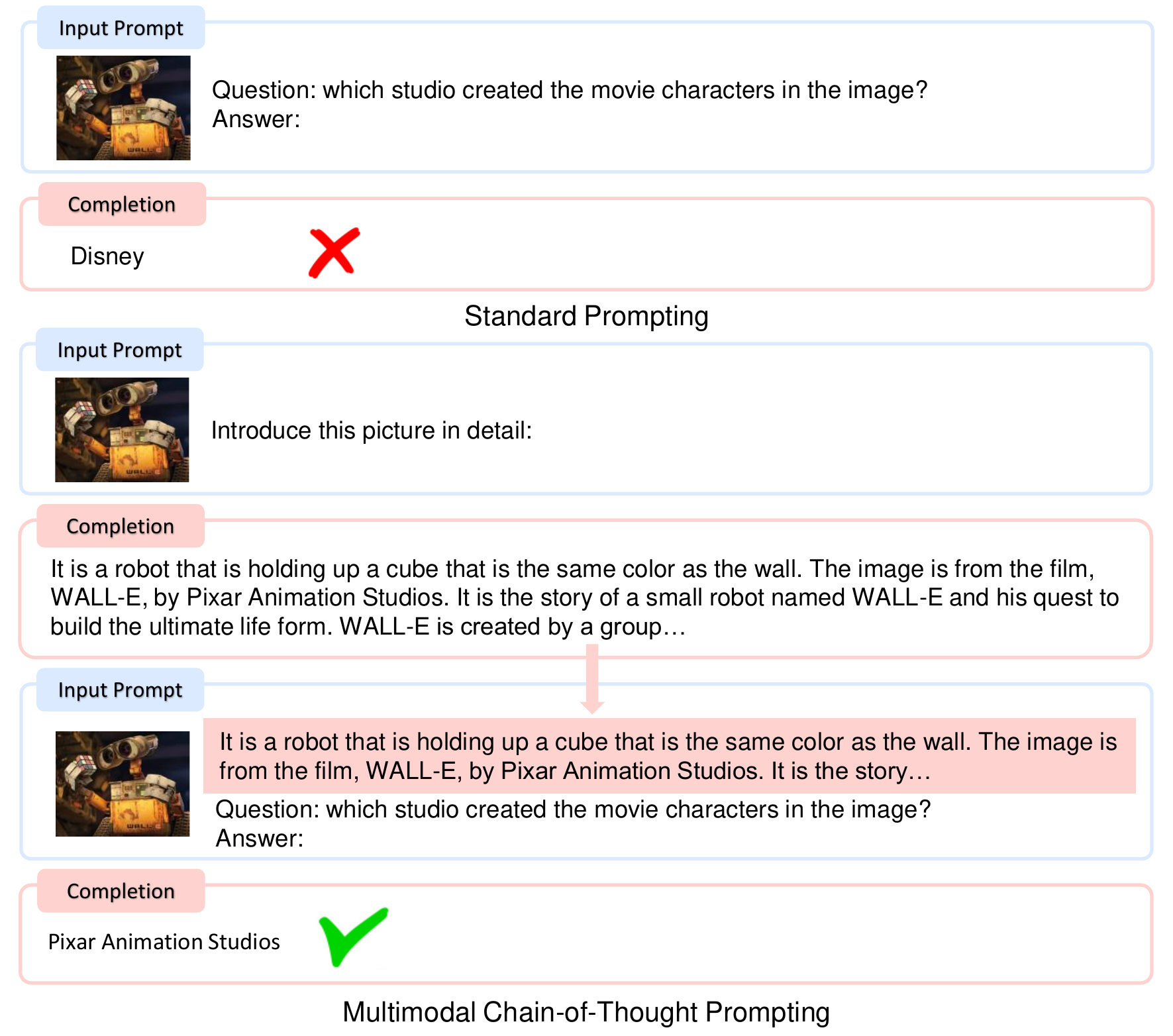}
\caption{
 Multimodal Chain-of-Thought prompting enables \our{} to generate a rationale first, then to tackle complex question-answering and reasoning tasks.
}
\label{fig:vl:mcot}
\end{figure*}

Chain-of-thought prompting~\cite{cot} allows large language models to generate a series of reasoning steps and decompose a multi-step problem into intermediate steps, which can significantly improve the performance in complex tasks. Motivated by chain-of-thought prompting, we investigate a multimodal chain-of-thought prompting using \our{}. As illustrated in Figure~\ref{fig:vl:mcot}, we break down perception-language tasks into two steps.
In the first stage, given an image, we use a prompt to guide the model to generate a rationale. The model is then fed the rationale and a task-aware prompt to produce the final results.

\subsubsection{Evaluation Setup}

We evaluate the ability of multimodal chain-of-thought prompting on the Rendered SST-2. 
We use the prompt \textit{\color{bluecode}``Introduce this picture in detail:''} to generate the content in the picture as the rationale. Then, we use the prompt  \textit{\color{bluecode}``\{rationale\} Question: what is the sentiment of the opinion? Answer: \{answer\}''} to predict the sentiment, where the answer is either positive or negative.

\subsubsection{Results}

We conduct experiments to evaluate the performance of the multimodal chain-of-thought prompting. Table~\ref{tbl:vl:mcot-ocr} shows that multimodal chain-of-thought prompting achieves a score of 72.9, which is 5.8 points higher than the standard prompting. By generating intermediate content, the model can recognize the text in the images and infer the sentiment of the sentences more correctly.


\begin{table}[ht]
\centering
\begin{tabular}{lc}
\toprule
\bf Models & \bf Accuracy \\
\midrule
CLIP ViT-B/32 & 59.6 \\
CLIP ViT-B/16 & 59.8 \\
CLIP ViT-L/14 & 64.0 \\
\our{} & 67.1 \\
~~~~w/ multimodal CoT prompting & \textbf{72.9} \\
\bottomrule
\end{tabular}
\vspace{0.2cm}
\caption{Multimodal chain-of-thought (CoT) prompting on Rendered SST-2 task.}
\label{tbl:vl:mcot-ocr}
\end{table}

\subsection{Zero-Shot Image Classification}
\label{sec:eval:image:zs_class}

We report the zero-shot image classification performance on ImageNet~\cite{imagenet}.
Image classification comprehends an entire image as a whole and aims to assign a label to the image.
We map each label to its category name in natural language.
The model is prompted to predict the category name to perform zero-shot image classification.

\subsubsection{Evaluation Setup}

Given an input image, we concatenate the image with the prompt \textit{\color{bluecode}``The photo of the''}.
The input is then fed into the model to obtain the category name of the image.
We evaluate the model on ImageNet~\cite{imagenet}, which contains 1.28M training images and 50k validation images in 1k object categories.
The prediction is classified as correct if it is exactly the same as the ground-truth category name.
The image resolution used for evaluation is 224$\times$224.
We use beam search to generate the category names and the beam size is 2.

\subsubsection{Results}

As shown in Table~\ref{tbl:vl:zs_in1k}, we report zero-shot results in both constrained and unconstrained settings.
The difference between the two settings is whether we use the 1k object category names to constrain the decoding.
\our{} significantly outperforms GIT~\cite{git} by 4.6\% under the constrained setting and 2.1\% under the unconstrained setting.

\begin{table}[ht]
\centering
\begin{tabular}{l cc}
\toprule
\textbf{Model} & \bf Without Constraints & \bf With Constraints \\
\midrule
GIT~\cite{git}          & 1.9   & 33.5          \\
\our{}       & \textbf{4.0} & \textbf{38.1} \\
\bottomrule
\end{tabular}
\vspace{0.2cm}
\caption{
Zero-shot image classification on ImageNet.
For the results with constraints, we use the 1k ImageNet object category names for constrained decoding. We report top-1 accuracy scores.
}
\label{tbl:vl:zs_in1k}
\end{table}

\subsection{Zero-Shot Image Classification with Descriptions}
\label{sec:eval:image:class}

\begin{figure*}[t]
\centering
\includegraphics[width=\textwidth]{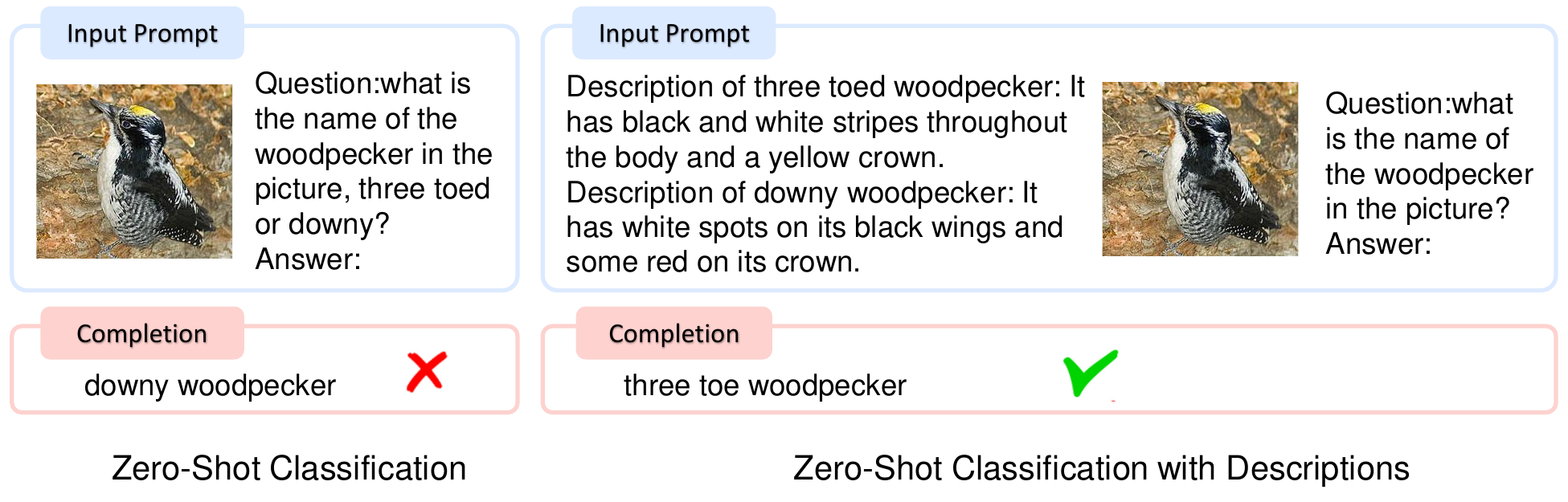}
\caption{
In-context verbal descriptions can help \our{} recognize visual categories better.
}
\label{fig:vl:incontext_classification}
\end{figure*}

The standard approach of image classification as above is to prompt the model for the specific name of the object depicted in the image. 
However, there are also some classification rules customized for different users and scenarios, such as the refined classification of complex animal subspecies.
We can utilize natural language descriptions to guide \our{} to distinguish images in the zero-shot setting, which makes the decision process more interpretable.

\subsubsection{Evaluation Setup}

Following CUB~\cite{CUB:dataset}, we construct a bird classification dataset that contains images and natural-language descriptions of categories.
The dataset has three groups of binary image classification.
Each group contains two animal categories with similar appearances.
Our goal is to classify images given the categories' descriptions.
Table~\ref{tbl:image_desc} presents the data samples.
The first group is from \cite{CUB:dataset}, while the other two groups are collected from the website.
Each category contains twenty images.

The evaluation procedure is illustrated in Figure~\ref{fig:vl:incontext_classification}.
For the zero-shot setting, we provide detailed descriptions of two specific categories and use the template \textit{\color{bluecode}``Question:what is the name of \{general category\} in the picture? Answer:''} to prompt the model for the specific category name in an open-ended manner.
To evaluate the effect of providing verbal descriptions in context, we also implement a zero-shot baseline without prompting descriptions. Instead, we provide the corresponding specific names in the prompt.

\subsubsection{Results}

\begin{table}[t!]
\centering
\setlength\tabcolsep{2.5pt}
\begin{tabular}{clcl}
\toprule
\multicolumn{2}{c}{\bf Category 1}  &  \multicolumn{2}{c}{\bf Category 2}   \\
\midrule
\multicolumn{2}{c}{three toed woodpecker}  &  \multicolumn{2}{c}{downy woodpecker} \\
\midrule
\multicolumn{1}{m{0.1\textwidth}}{\includegraphics[width=.1\textwidth]{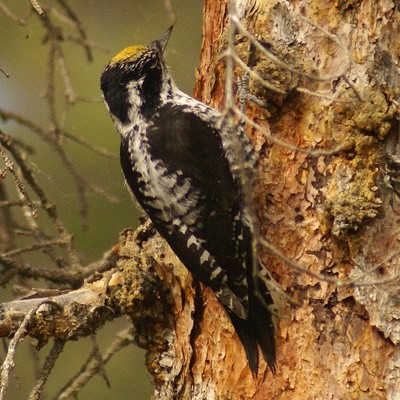}} & \multicolumn{1}{m{0.35\textwidth}}{It has black and white stripes throughout the body and a yellow crown.} & \multicolumn{1}{m{0.1\textwidth}}{\includegraphics[width=.1\textwidth]{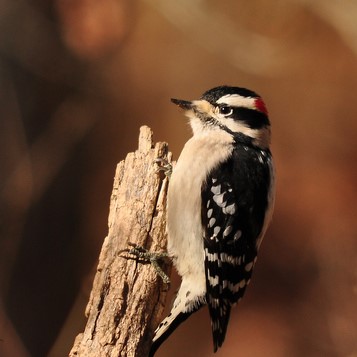}} & \multicolumn{1}{m{0.35\textwidth}}{It has white spots on its black wings and some red on its crown.} \\
\midrule
\multicolumn{2}{c}{Gentoo penguin}  &  \multicolumn{2}{c}{royal penguin}\\
\midrule
\multicolumn{1}{m{0.1\textwidth}}{\includegraphics[width=.1\textwidth]{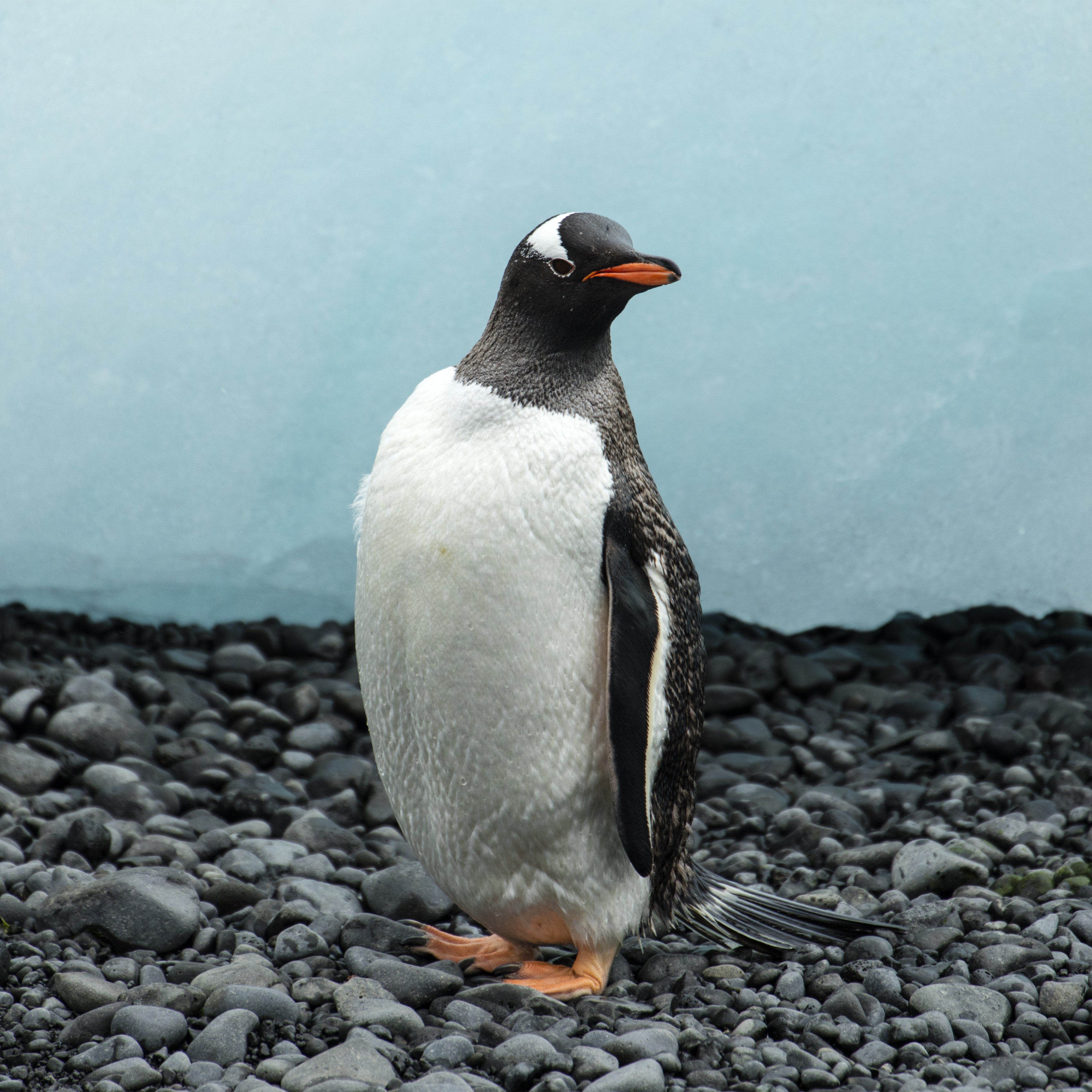}} & \multicolumn{1}{m{0.35\textwidth}}{It has a black head and white patch above its eyes.} & \multicolumn{1}{m{0.1\textwidth}}{\includegraphics[width=.1\textwidth]{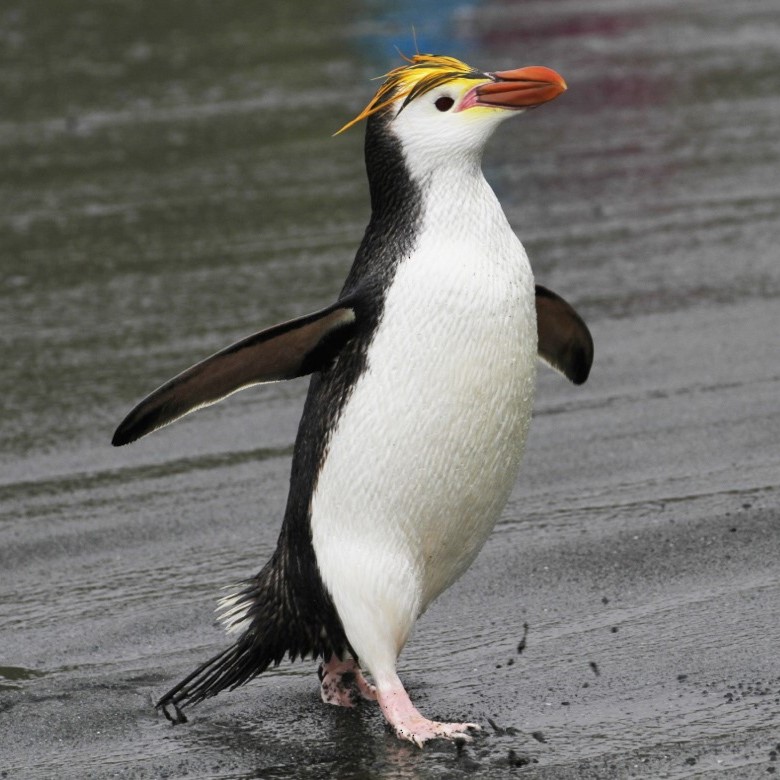}} & \multicolumn{1}{m{0.35\textwidth}}{It has a white face and a yellow crown.} \\
\midrule
\multicolumn{2}{c}{black throated sparrow}  &  \multicolumn{2}{c}{fox sparrow} \\
\midrule
\multicolumn{1}{m{0.1\textwidth}}{\includegraphics[width=.1\textwidth]{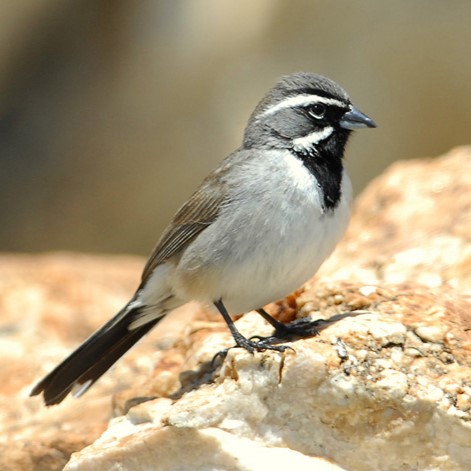}} & \multicolumn{1}{m{0.35\textwidth}}{It has white underparts and a distinctive black bib on the throat.} & \multicolumn{1}{m{0.1\textwidth}}{\includegraphics[width=.1\textwidth]{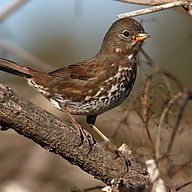}} & \multicolumn{1}{m{0.35\textwidth}}{It has a reddish-brown plumage and a streaked breast.} \\
\bottomrule
\end{tabular}
\vspace{0.2cm}
\caption{The detailed descriptions of different categories for in-context image classification.}
\label{tbl:image_desc}
\end{table}

The evaluation results are shown in Table~\ref{tbl:image_desc_acc}.
We observe that providing descriptions in context can significantly improve the accuracy of image classification.
The consistent improvements indicate that \our{} can perceive the intentions of instructions and well align the concepts in language modality with visual features in vision modality.

\begin{table}[ht]
\centering
\begin{tabular}{lc}
\toprule
\bf Settings & \bf Accuracy \\
\midrule
Without Descriptions & 61.7 \\
With Descriptions & \textbf{90.0} \\
\bottomrule
\end{tabular}
\vspace{0.2cm}
\caption{Results of zero-shot image classification without and with verbal descriptions.}
\label{tbl:image_desc_acc}
\end{table}

\subsection{Language Tasks}
\label{sec:eval:language}

The models are evaluated on the language tasks given task instructions (i.e., zero-shot) or several demonstration examples (i.e., few-shot).
Text inputs are directly fed into the models as in vanilla language models.

\subsubsection{Evaluation Setup}

We train a language model (\tlm{}) baseline with the same text corpora and training setup.
We evaluate \our{} and the \tlm{} baseline on eight language tasks, including cloze and completion tasks (i.e, StoryCloze, HellaSwag), Winograd-style tasks (i.e, Winograd, Winogrande), commonsense reasoning (i.e, PIQA), and three datasets BoolQ, CB, and COPA from the SuperGLUE benchmark~\cite{superglue}.
The detailed descriptions of these datasets are provided in Appendix~\ref{app:corpora:data:lang}. We conduct experiments under zero-shot and few-shot settings. We evaluate each test example by randomly sampling examples from the training set as demonstrations. We set the number of shots to 0, 1, and 4 in our experiments.

\subsubsection{Results}

Table~\ref{tbl:lang:few_shot} presents the in-context learning performance of language tasks. \our{} achieves comparable or even better performance in cloze completion and commonsense reasoning tasks when compared to \tlm{}. In terms of the average result across all these datasets, \tlm{} performs better in zero-shot and one-shot settings, whereas our model performs better in few-shot ($k=4$) settings.
The results indicate that \our{} also handles language-only tasks well and achieves favorable performance across datasets.
In addition, Section~\ref{sec:eval:visual_commonsense} shows that MLLMs learn better visual commonsense knowledge compared with LLMs.

\begin{table}[ht]
\centering
\begin{tabular}{l c c c c c c}
\toprule
\multirow{2}*{\textbf{Task}}
 & \multicolumn{2}{c}{\textbf{Zero-shot}} & \multicolumn{2}{c}{\textbf{One-shot}} & \multicolumn{2}{c}{\textbf{Few-shot} ($k=4$)} \\
 \cmidrule(r){2-3} \cmidrule(l){4-5} \cmidrule(l){6-7}
 & LLM & \our{} & LLM & \our{} & LLM & \our{} \\
 \midrule

StoryCloze & \textbf{72.9} & 72.1          & \textbf{72.9} & 72.2          & \textbf{73.1} & 72.3          \\
HellaSwag  & \textbf{50.4} & 50.0          & \textbf{50.2} & 50.0          & \textbf{50.4} & 50.3          \\[0.2cm]
Winograd   & \textbf{71.6} & 69.8          & \textbf{71.2} & 68.4          & \textbf{70.9} & 69.8          \\
Winogrande & \textbf{56.7} & 54.8 & \textbf{56.7} & 54.5          & \textbf{57.0} & 55.7          \\[0.2cm]
PIQA       & \textbf{73.2} & 72.9          & \textbf{73.0} & 72.5 & \textbf{72.6} & 72.3          \\[0.2cm]
BoolQ      & \textbf{56.4} & \textbf{56.4} & 55.1          & \textbf{57.2} & 58.7          & \textbf{59.2} \\
CB         & 39.3          & \textbf{44.6} & 41.1          & \textbf{48.2} & 42.9          & \textbf{53.6} \\
COPA       & \textbf{68.0} & 63.0          & \textbf{69.0} & 64.0          & \textbf{69.0} & 64.0          \\
\midrule
\textbf{Average} & {61.1} & 60.5          & {61.2} & 60.9          & 61.8          & {62.2} \\
           
\bottomrule 
\end{tabular}
\vspace{0.2cm}
\caption{Performance comparisons of language tasks between \our{} and LLM. We use the same textual data and training setup to reimplement a language model.
Both models do not use instruction tuning for fair comparisons.
}
\label{tbl:lang:few_shot}
\end{table}

\subsection{Cross-modal Transfer}
\label{sec:eval:transfer}
Cross-modal transferability allows a model to learn from one modality (such as text, image, audio, etc.) and transfer the knowledge to the other modalities. This skill can enable a model to perform various tasks across different modalities. In this part, we evaluate the cross-model transferability of \our{} on several benchmarks.

\subsubsection{Transfer from Language to Multimodal: Language-Only Instruction Tuning}
\label{sec:eval:instruct}

To evaluate the effect of language-only instruction tuning, we conduct an ablation study using four datasets: COCO, Flickr30k, VQAv2, and VizWiz. These datasets consist of image captioning and visual questions anwsering. The evaluation metrics are: CIDEr scores for COCO/Flickr30k and VQA accuracy for VQAv2/VizWiz.

Table~\ref{tbl:vl:instruct} shows the experimental results. Language-only instruction tuning boosts our model's performance by 1.9 points on Flickr30k, 4.3 points on VQAv2, and 1.3 points on VizWiz. Our experiments show that language-only instruction tuning can significantly improve the model's instruction-following capabilities across modalities.
The results also indicate that our model can transfer the instruction-following capability from language to other modalities.

\begin{table}[ht]
\centering
\begin{tabular}{l cccc}
\toprule
\textbf{Model} & \textbf{COCO} & \textbf{Flickr30k} & \textbf{VQAv2} & \textbf{VizWiz} \\
\midrule
\our{}           & 84.7 &  \textbf{67.1}  & \textbf{51.0}	& \textbf{29.2} \\
~ w/o \lait{} & \textbf{87.6} &  65.2  & 46.7	&  27.9	\\
\bottomrule
\end{tabular}
\vspace{0.2cm}
\caption{
Ablation study on language-only instruction tuning.
We report CIDEr scores for COCO and Flickr30k, and VQA
accuracy scores for VQAv2 and VizWiz.
}
\label{tbl:vl:instruct}
\end{table}

\subsubsection{Transfer from Multimodal to Language: Visual Commonsense Reasoning}
\label{sec:eval:visual_commonsense}

Visual commonsense reasoning tasks require an understanding of the properties of everyday objects in the real world, such as color, size, and shape. These tasks are challenging for language models because they may require more information about object properties than what is available in texts. To investigate the visual commonsense capabilities, we compare the zero-shot performance of \our{} and \tlm{} on visual commonsense reasoning tasks.

\paragraph{Evaluation Setup}
We compare \our{} and the \tlm{} baseline on three object commonsense reasoning datasets, \textsc{RelativeSize}~\cite{Bagherinezhad2016AreEB}, \textsc{MemoryColor}~\cite{Norlund2021TransferringKF} and \textsc{ColorTerms}~\cite{Bruni2012DistributionalSI} datasets. Table~\ref{tbl:vl:objectreasoning:data} shows some examples of object size and color reasoning tasks. \textsc{RelativeSize} contains 486 object pairs from 41 physical objects. The model is required to predict the size relation between two objects in a binary question-answering format with ``Yes''/``No'' answers. \textsc{MemoryColor} and \textsc{ColorTerms} require the model to predict the color of objects from a set of 11 color labels in a multiple-choice format. We use only text as our input and do not include any images. We measure the accuracy of our model on these three datasets.

\begin{table*}[ht]
\centering
\small
\begin{tabular}{@{}l ll l}
\toprule
\bf Task &\bf  Example Prompt & \bf Object / Pair &\bf  Answer  \\
\midrule
{Object Size Reasoning} &   \textit{\color{bluecode}Is \{Item1\} larger than \{Item2\}? \{Answer\} } & (\textit{sofa}, \textit{cat}) & \textit{Yes} \\
{Object Color Reasoning} &  \textit{\color{bluecode} The color of \{Object\} is? \{Answer\} }  & \textit{the sky} & \textit{blue} \\
\bottomrule
\end{tabular}
\caption{Evaluation examples of object size and color reasoning.}
\label{tbl:vl:objectreasoning:data}
\end{table*}

\paragraph{Results}

Table~\ref{tbl:vl:objectreasoning} presents the zero-shot performance of \our{} and \tlm{} on visual commonsense reasoning tasks. \our{} significantly outperforms \tlm{} by 1.5\% on \textsc{RelativeSize}, 14.7\% on \textsc{MemoryColor}, and 9.7\% on \textsc{ColorTerms} dataset. The consistent improvements indicate that \our{} benefits from the visual knowledge to complete the corresponding visual commonsense reasoning.
The reason for \our{}'s superior performance is that it has modality transferability, which enables the model to transfer visual knowledge to language tasks. On the contrary, \tlm{} has to rely on textual knowledge and clues to answer visual commonsense questions, which limits its ability to reason about object properties.

\begin{table}[ht]
\centering
\begin{tabular}{@{}lccc@{}}
\toprule
\multirow{2}{*}{\textbf{Model}} & \textbf{Size Reasoning} & \multicolumn{2}{c}{\textbf{Color Reasoning} }\\
 & {\textsc{RelativeSize}} & {\textsc{MemoryColor}} & {\textsc{ColorTerms}} \\
\midrule
\multicolumn{4}{l}{~\textit{Using retrieved images}} \\
\textsc{VaLM}~\cite{valm} & 85.0 & 58.6 & 52.7 \\
\midrule
\multicolumn{4}{l}{~\textit{Language-only zero-shot evaluation}} \\
\tlm{}              & 92.7	& 61.4	&   63.4   \\
\our{}              & \textbf{94.2}	& \textbf{76.1}	&   \textbf{73.1}   \\
\bottomrule
\end{tabular}
\vspace{0.2cm}
\caption{
Zero-shot visual commonsense reasoning on \textsc{RelativeSize}, \textsc{MemoryColor}, and \textsc{ColorTerms} datasets.
Accuracy scores are reported.
}
\label{tbl:vl:objectreasoning}
\end{table}

\section{Conclusion}

In this work, we introduce \our{}, a multimodal large language model that can perceive general modalities, follow instructions, and perform in-context learning.
The models trained on web-scale multimodal corpora achieve promising results across a wide range of language tasks and multimodal tasks.
We show that going from LLMs to MLLMs enables new capabilities and opportunities.
In the future, we would like to scale up \our{} in terms of model size~\cite{torchscale,magneto,xmoe}, and integrate the speech~\cite{valle} capability into \our{}.
In addition, \our{} can be used as a unified interface for multimodal learning, e.g., enabling using instructions and examples to control text-to-image generation.

\bibliographystyle{alpha}
\bibliography{kosmos}

\nocite{cm3}
\nocite{cm3-retrieval}
\nocite{frozen}
\nocite{blip2}
\nocite{FROMAGe}

\newpage
\appendix

\section{Hyperparameters}
\label{app:hyperparam}
\subsection{Training}
We report the detailed model hyperparameter settings of \our{} in Table~\ref{tbl:hyperparam:vl:pt:model} and training hyperparameters in Table~\ref{tbl:hyperparam:vl:pt:opt}.

\begin{table}[ht]
\centering
\begin{tabular}{lc}
\toprule
\textbf{Hyperparameters} & \\ \midrule
Number of layers            & 24    \\
Hidden size                 & 2,048 \\
FFN inner hidden size       & 8,192 \\
Attention heads             & 32 \\
Dropout                     & 0.1 \\
Attention dropout           & 0.1 \\
Activation function         & GeLU~\cite{gelu} \\
Vocabulary size             & 64,007 \\
Soft tokens $V$ size        & 64 \\
Max length                  & 2,048 \\
Relative position embedding & xPos~\cite{xpos} \\
Initialization              & Magneto~\cite{magneto} \\
\bottomrule
\end{tabular}
\vspace{0.2cm}
\caption{Hyperparameters of causal language model of \our{}}
\label{tbl:hyperparam:vl:pt:model}
\end{table}

\begin{table}[ht]
\centering
\begin{tabular}{lc}
\toprule
\textbf{Hyperparameters} & \\ \midrule
Training steps                     &       300,000 \\
Warmup steps                      &       375 \\
Batch size of text corpora         &       256 \\
Max length of text corpora         &       2,048 \\
Batch size of image-caption pairs  &       6,144 \\
Batch size of interleaved data   &       128 \\
Optimizer & Adam \\
Learning rate & 2e-4 \\
Learning Rate Decay & Linear \\
Adam $\epsilon$ & 1e-6 \\
Adam $\beta$ & (0.9, 0.98) \\
Weight decay & 0.01 \\
\bottomrule
\end{tabular}
\vspace{0.2cm}
\caption{Training hyperparameters of \our{}}
\label{tbl:hyperparam:vl:pt:opt}
\end{table}

\subsection{Language-Only Instruction Tuning}
The detailed instruction tuning hyperparameters are listed in Table~\ref{tbl:hyperparam:vl:instruct:opt}.
\label{app:hyperparam:inst}
\begin{table}[ht]
\centering
\begin{tabular}{lc}
\toprule
\textbf{Hyperparameters} & \\ \midrule
Training steps                     &       10,000 \\
Warmup steps                      &       375 \\
Batch size of instruction data     &       256 \\
Batch size of text corpora         &       32 \\
Batch size of image-caption pairs  &       768 \\
Batch size of interleaved data   &       16 \\
Learning rate & 2e-5 \\
\bottomrule
\end{tabular}
\vspace{0.2cm}
\caption{Instruction tuning hyperparameters of \our{}}
\label{tbl:hyperparam:vl:instruct:opt}
\end{table}




\section{Datasets}
\label{app:corpora}
\subsection{Pretraning}
\label{app:corpora:data:pt}

\subsubsection{Text Corpora}
\label{app:corpora:data:pt:lang}

\our{} is trained on The Pile~\cite{pile} and Common Crawl. The Pile is an 800 GB English text corpus combining 22 diverse sources. We select a subset with seven sources from The Pile. Common Crawl is also included in training corpora. Common Crawl takes snapshots of the web, which contains massive amounts of language data. Table~\ref{tbl:lang:pt} provides a full overview of the language datasets that were used in the training of \our{} model.
These data sources can be divided into the following three categories:
\begin{itemize}[leftmargin=*]
\item \textbf{Academic}: NIH Exporter
\item \textbf{Internet}: Pile-CC, OpenWebText2, Wikipedia (English), CC-2020-50, CC-2021-04, Realnews
\item \textbf{Prose}: BookCorpus2, Books3, Gutenberg~\cite{pg19}, CC-Stories
\end{itemize}

\begin{table}[ht]
\centering
\begin{tabular}{l c c c c c c}
\toprule
\textbf{Datasets} & \textbf{Tokens (billion)} & \textbf{Weight (\%)} & \textbf{Epochs} \\
 \midrule
OpenWebText2      & 14.8                      & 21.8\%               & 1.47             \\
CC-2021-04        & 82.6                      & 17.7\%               & 0.21             \\
Books3            & 25.7                      & 16.2\%               & 0.63             \\
CC-2020-50        & 68.7                      & 14.7\%               & 0.21             \\
Pile-CC           & 49.8                      & 10.6\%               & 0.21             \\
Realnews          & 21.9                      & 10.2\%               & 0.46             \\
Wikipedia         & 4.2                       & 5.4\%                & 1.29             \\
BookCorpus2       & 1.5                       & 1.1\%                & 0.75             \\
Gutenberg (PG-19) & 2.7                       & 1.0\%                & 0.38             \\
CC-Stories        & 5.3                       & 1.0\%                & 0.19             \\
NIH ExPorter      & 0.3                       & 0.2\%                & 0.75            \\
\bottomrule 
\end{tabular}
\vspace{0.2cm}
\caption{Language datasets used to train the \our{} model.}
\label{tbl:lang:pt}
\end{table}

\subsubsection{Image-Caption Pairs}
\label{app:corpora:data:pt:image-caption}

\our{} is trained on image-caption pairs constructed from several datasets, including English LAION-2B~\cite{laion5b}, LAION-400M~\cite{laion400m}, COYO-700M~\cite{coyo700m} and Conceptual Captions~\cite{cc3m,cc12m}.
LAION-2B, LAION-400M, and COYO-700M datasets are extracted by parsing out image URLs and alt-texts of web pages from the Common Crawl web data.
LAION-2B contains about 2B English image-caption pairs, LAION-400M consists of 400M English image-caption pairs, and COYO-700M has 700M English image-caption pairs.
Conceptual Captions contains 15M English image-caption pairs and consists of two datasets: CC3M and CC12M, which are also collected from internet webpages using a Flume pipeline.
For Conceptual Captions, we discard pairs whose captions contain special tags such as ``<PERSON>''.

\subsubsection{Interleaved Data}
\label{app:corpora:data:pt:multimodal}

We collect a large corpus of 2 billion web pages from the snapshots of common crawls. To ensure quality and relevance, we apply several filtering criteria. First, we discard any pages that are not written in English. Second, we discard any pages that do not have images interspersed in the text. Third, we discard any images that have a resolution lower than 64 by 64 pixels or that are single-colored. Fourth, we discard any text that is not meaningful or coherent, such as spam or gibberish. We use some heuristics to identify and remove gibberish text containing emoji symbols, hashtags, and URL links. After applying these filters, we end up with about 71 million documents for training.

\subsection{Data Format}
\label{app:corpora:data:format}

The training data is organized in the format as follows:
\begin{table}[ht]
\centering
\begin{tabular}{l p{10.5cm}}
\toprule
\textbf{Datasets} & \textbf{Format Examples}  \\
 \midrule
\textbf{Text}      & \texttt{<s>} \our{} can perceive multimodal input, learn in context, and generate output. \texttt{</s>}           \\
\textbf{Image-Caption}        & \texttt{<s>} \texttt{<image>} Image Embedding \texttt{</image>} WALL-E giving potted plant to EVE. \texttt{</s>}           \\
\textbf{Multimodal}        & \texttt{<s>} \texttt{<image>} Image Embedding \texttt{</image>} This is WALL-E. \texttt{<image>} Image Embedding \texttt{</image>} This is EVE. \texttt{</s>}        \\
\bottomrule 
\end{tabular}
\vspace{0.2cm}
\caption{The examples of the data format to train the \our{} model.}
\label{tbl:data:format}
\end{table}

\section{Evaluation}

\subsection{Input Format Used for Perception-Language Tasks}

\begin{figure*}[ht]
\centering
\includegraphics[width=0.9\textwidth]{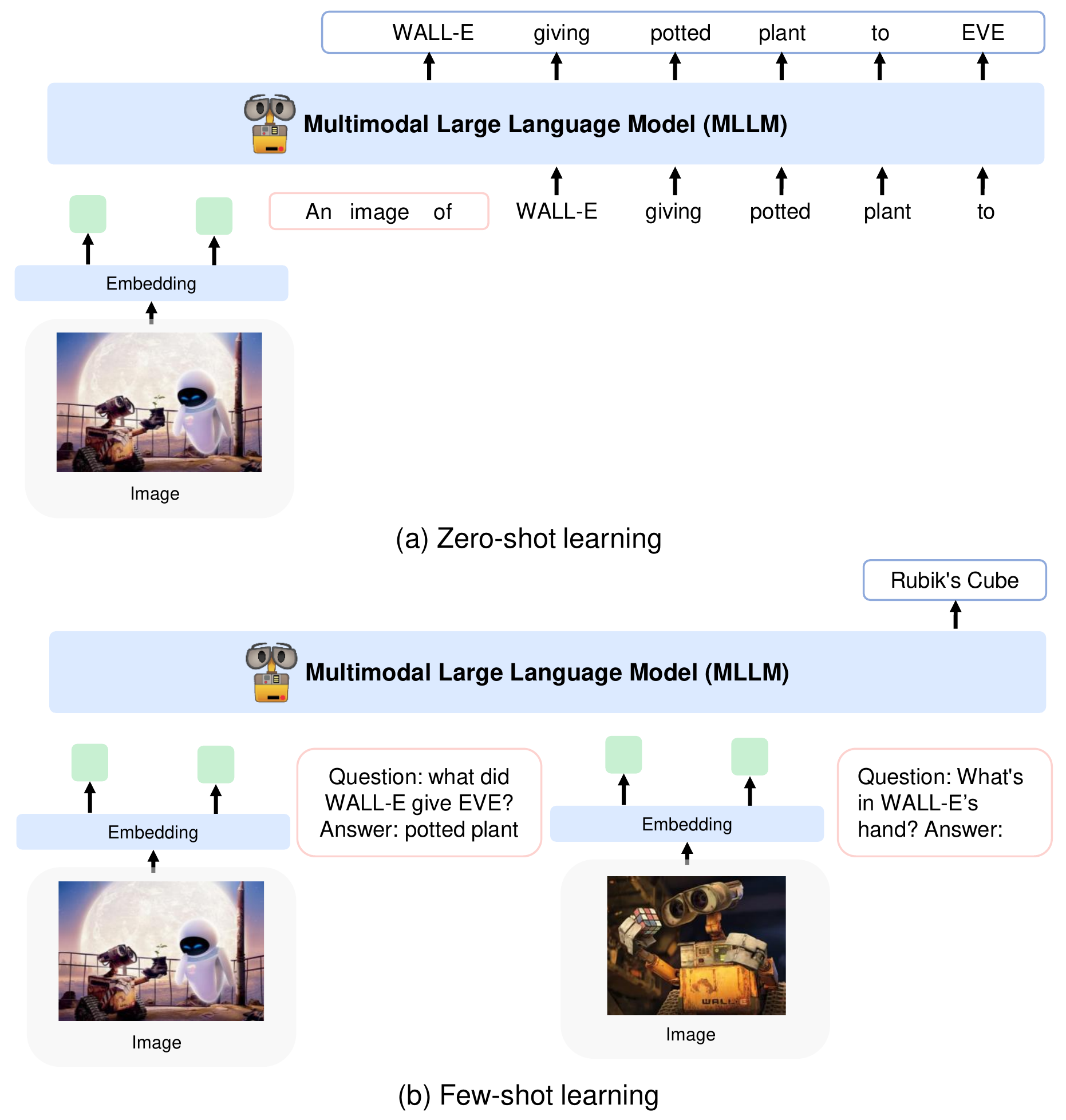}
\caption{We evaluate \our{} on the perception-language tasks  in zero- and few-shot settings.
(a) Zero-shot learning, e.g., zero-shot image captioning with language prompts.
(b) Few-shot learning, e.g., visual question answering with in-context learning. 
}
\label{fig:vl:eval:protocol}
\end{figure*}

Figure~\ref{fig:vl:eval:protocol} shows how we conduct zero-shot and few-shot evaluations on perception-language tasks.

\subsection{Language Tasks}
\label{app:corpora:data:lang}

We conduct experiments on language tasks in four categories:
\begin{itemize}[leftmargin=*]
\item Cloze and completion tasks: StoryCloze~\cite{storycloze}, HellaSwag~\cite{hellaswag}
\item Winograd-style tasks: Winograd~\cite{wsc}, Winogrande~\cite{winogrande}
\item Commonsense reasoning: PIQA~\cite{piqa}
\item Three datasets from SuperGLUE benchmark~\cite{superglue}: BoolQ~\cite{boolq}, CB~\cite{cb}, COPA~\cite{copa}
\end{itemize}

\subsection{WebSRC Task Examples}
\label{app:corpora:data:vl:websrc}

\begin{figure}[h]
     \centering
     \begin{subfigure}[b]{0.3\textwidth}
         \centering
         \includegraphics[width=\textwidth]{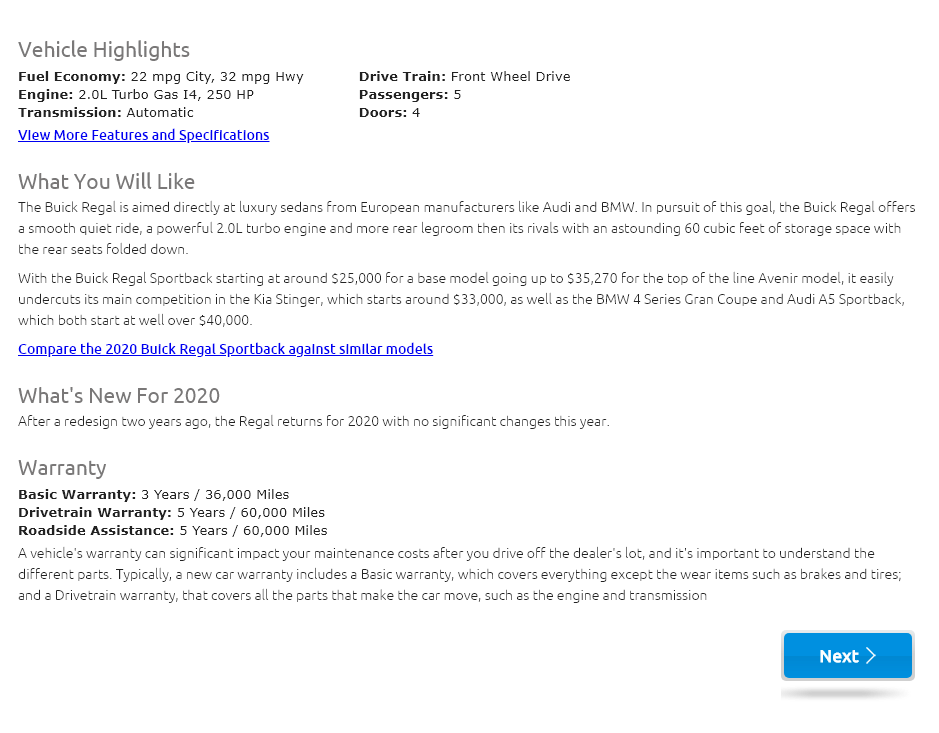}
         \caption{Question is ``\textit{What is the type of this drive?}''}
         \label{fig:y equals x}
     \end{subfigure}
     \hfill
     \begin{subfigure}[b]{0.6\textwidth}
         \centering
         \includegraphics[width=\textwidth]{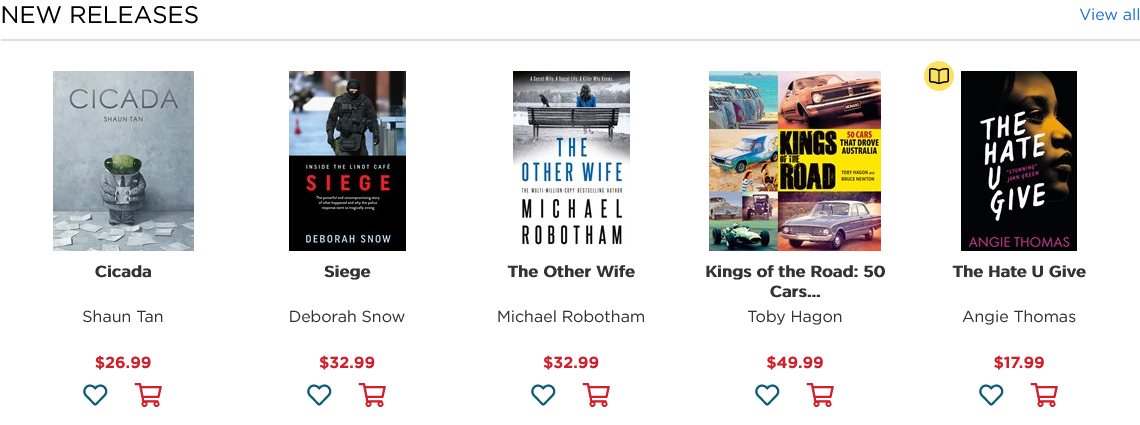}
         \caption{Question is ``\textit{Who is the author of "Cicada"?}''}
     \end{subfigure}
     \hfill
        \caption{Examples form WebSRC~\cite{websrc}.}
        \label{fig:three graphs}
\end{figure}

\end{document}